\documentclass[final]{cvpr}

\usepackage{amsmath,amsfonts,bm}

\def\eqref#1{(\ref{#1})}
\def\1{\bm{1}}

\DeclareMathAlphabet{\mathsfit}{\encodingdefault}{\sfdefault}{m}{sl}
\SetMathAlphabet{\mathsfit}{bold}{\encodingdefault}{\sfdefault}{bx}{n}

\usepackage{times}
\usepackage{epsfig}
\usepackage{graphicx}
\usepackage{amsmath}
\usepackage{amssymb}
\usepackage{multirow}
\usepackage{footmisc}
\usepackage{color}
\usepackage[dvipsnames]{xcolor}
\usepackage{colortbl,booktabs}
\usepackage{threeparttable}
\usepackage{nopageno}
\usepackage{comment}
\usepackage{url}
\usepackage{algorithm}
\usepackage{algpseudocode}
\usepackage{mathtools}
\usepackage{pifont}
\usepackage{lipsum}
\usepackage{color,xcolor}
\definecolor{Tianlong_color}{rgb}{0.858, 0.188, 0.478}
\usepackage{adjustbox}
\usepackage{booktabs}       % professional-quality tables
\usepackage{amsfonts}       % blackboard math symbols
\usepackage{nicefrac}       % compact symbols for 1/2, etc.
\usepackage{microtype}      % microtypography
\usepackage{makecell}
\usepackage{diagbox}

\DeclarePairedDelimiterX{\inp}[2]{\langle}{\rangle}{#1, #2}

\DeclareMathAlphabet\mathbfcal{OMS}{cmsy}{b}{n}

\usepackage[pagebackref=true,breaklinks=true,colorlinks,bookmarks=false]{hyperref}
\usepackage[pagebackref=true,breaklinks=true,colorlinks,bookmarks=false]{hyperref}

\def\cvprPaperID{6316} % *** Enter the CVPR Paper ID here
\def\confYear{CVPR 2021}

\begin{document}

%%%%%%%%% TITLE
\title{The Lottery Tickets Hypothesis for Supervised and Self-supervised \\ Pre-training in Computer Vision Models}

\author{%
  Tianlong Chen\textsuperscript{1}, Jonathan Frankle\textsuperscript{2}, Shiyu Chang\textsuperscript{3}, Sijia Liu\textsuperscript{3,4}, Yang Zhang\textsuperscript{3}, \\ Michael Carbin\textsuperscript{2}, Zhangyang Wang\textsuperscript{1}\\
  \textsuperscript{1}University of Texas at Austin, \textsuperscript{2}MIT CSAIL
  \textsuperscript{3}MIT-IBM Watson AI Lab, \textsuperscript{4}Michigan State University \\
  \small{\texttt{\{tianlong.chen,atlaswang\}@utexas.edu,\{jfrankle,mcarbin\}@csail.mit.edu,}} \\
  \small{\texttt{\{shiyu.chang,yang.zhang2\}@ibm.com,liusiji5@msu.ed}}
}

\maketitle

%%%%%%%%% ABSTRACT
\begin{abstract}
\vspace{-1em}
The computer vision world has been re-gaining enthusiasm in various pre-trained models, including both classical ImageNet supervised pre-training and recently emerged self-supervised pre-training such as simCLR~\cite{chen2020simple} and MoCo~\cite{he2020momentum}. Pre-trained weights often boost a wide range of downstream tasks including classification, detection, and segmentation. Latest studies suggest that pre-training benefits from gigantic model capacity~\cite{chen2020big}. We are hereby curious and ask: after pre-training, does a pre-trained model indeed have to stay large for its downstream transferability? 

In this paper, we examine supervised and self-supervised pre-trained models through the lens of the \textit{lottery ticket hypothesis} (LTH) \cite{frankle2018the}. LTH identifies highly sparse matching subnetworks that can be trained in isolation from (nearly) scratch yet still reach the full models' performance. We extend the scope of LTH and  question whether matching subnetworks still exist in pre-trained computer vision models, that enjoy the same downstream transfer performance. Our extensive experiments convey an overall positive message: from all pre-trained weights obtained by ImageNet classification, simCLR, and MoCo, we are consistently able to locate such matching subnetworks at $59.04\%$ to $96.48\%$ sparsity that transfer \text{universally} to multiple downstream tasks, whose performance see no degradation compared to using full pre-trained weights. Further analyses reveal that subnetworks found from different pre-training tend to yield diverse mask structures and perturbation sensitivities. We conclude that the core LTH observations remain generally relevant in the pre-training paradigm of computer vision, but more delicate discussions are needed in some cases. Codes and pre-trained models will be made available at: \small{\url{https://github.com/VITA-Group/CV_LTH_Pre-training}}.

\vspace{-0.5em}
%from both self-supervised and supervised learning like simCLR \cite{chen2020simple}, MoCo \cite{he2020momentum} and ImageNet \textbf{pre-training}, have gained growing popularity to serve as the standard starting point for training on various downstream tasks, including classification, detection, and segmentation. In parallel, studies on the \textbf{lottery ticket hypothesis} show prevailing successes on CV and natural language processing in locating sparse \textbf{matching subnetworks} capable of training in isolation to full model performance. In this work, we bridge these researches to investigate whether such trainable, transferable subnetworks exist in both self-supervised and supervised pre-training. Unlike previous work in CV, we find subnetworks at pre-trained initialization rather than after some amount of training. Subnetworks found on the simCLR, MoCo and Imagenet pre-training task transfer \textbf{universally} to multiple downstream classification tasks, while they show a limited transferability for downstream detection and segmentation tasks. We also show that subnetworks found on different pre-training have diverse structures and are more sensitive to structural perturbations. Our comprehensive results demonstrate that the core lottery ticket observations maintain relevant in the prevalent paradigm of pre-training and transfer learning.
\end{abstract}

%%%%%%%%% BODY TEXT
\vspace{-1em}
\section{Introduction}

\begin{figure}[t] 
\centering
\vspace{-1.8em}
\includegraphics[width=1\linewidth]{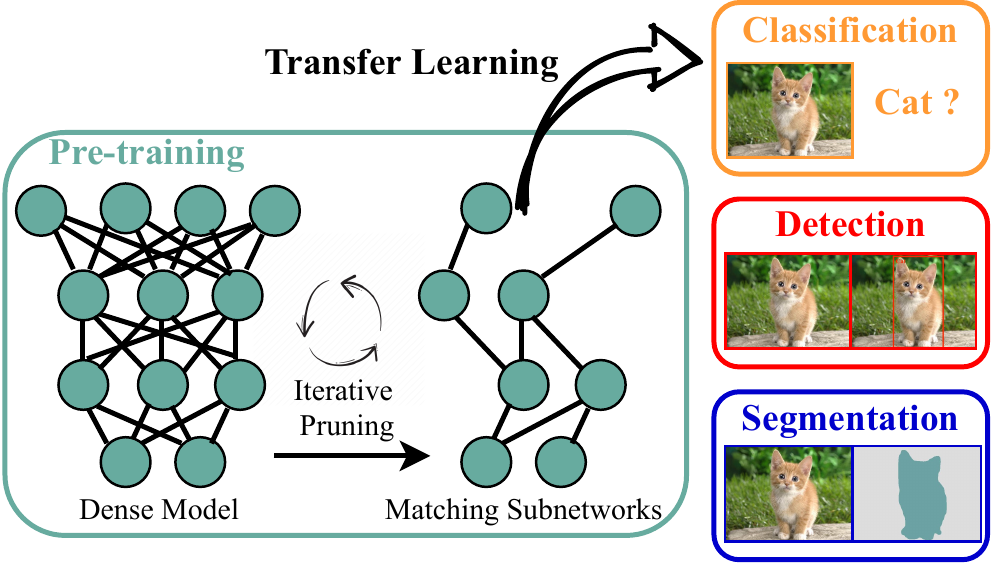}
\caption{Overview of our work paradigm: from pre-trained CV models (both supervised and self-supervised), we study the existence of matching subnetworks that are transferable to many downstream tasks, with little performance degradation compared to using full pre-trained weights. We find \textit{task-agnostic, universally transferable} subnetworks at pre-trained initialization, for \textbf{classification}, \textbf{detection}, and \textbf{segmentation} tasks. 
% while the \textbf{detection} task (in particular, two-stage object detector) seems to require \textit{task-specific} search of matching subnetworks.
}
\vspace{-1.5em}
\label{fig:framework}
\end{figure}

\vspace{-0.5em}
Deep neural networks pre-trained on large-scale datasets prevail as general-purpose feature extractors \cite{donahue2014decaf}. Moving beyond the most traditional greedy unsupervised pre-training \cite{bengio2006greedy}, the most popular pre-training in computer vision (CV) is arguably to train the model for supervised classification on ImageNet \cite{deng2009imagenet}. Such \textit{supervised pre-training} enables the network to learn a hierarchy of generalizable features \cite{huh2016makes}; it is widely acknowledged \cite{girshick2014rich} to not only benefit the subsequent fine-tuning on other visual classification datasets (especially in small datasets and few-shot learning \cite{russakovsky2015imagenet,tian2020rethinking}), but also to accelerate/improve the training for different, more complicated types of downstream vision tasks, such as object detection and semantic segmentation \cite{oquab2014learning,he2019rethinking}. 

Several state-of-the-art \textit{self-supervised pre-training}, such as simCLR \cite{chen2020simple,chen2020big} and MoCo \cite{he2020momentum,chen2020improved}, have demonstrated that it is instead possible to use unlabeled data in pre-training. Their methods refer to no actual labels in pre-training, but instead leverage self-generated pseudo labels \cite{doersch2015unsupervised,dosovitskiy2014discriminative} or contrasting augmented views \cite{chen2020simple}. Impressively, self-supervised pre-training yields pre-trained weights with comparable or even better transferability and generalization, for various downstream tasks, compared to their supervised pre-training counterparts.

A few recent efforts have shown to successfully \textit{scale up} pre-training in CV. That is perhaps most natural for self-supervised pre-training, since unlabeled images are cheap and easily accessible. Chen et al. \cite{chen2020big} investigated to boost simCLR with massive unlabeled data in a task-agnostic way, and pointed out the key ingredient to be the use of big (deep and wide) networks \textit{during pretraining and fine-tuning}. The authors found that, the fewer the labels, the more this approach (task-agnostic use of unlabeled data) benefits from a bigger network. \textit{After fine-tuning}, the big network is reduced into a much smaller one with little performance loss by using \textit{task-specific} distillation. We additionally note the latest works suggesting that supervised fine-tuning can also scale up to larger models and datasets beyond ImageNet \cite{dosovitskiy2020image}.

% The excessive capacity of pre-trained CV models might rightfully remind of the mainstream scheme in natural language processing (NLP) \cite{devlin2018bert,radford2019language}: one first pre-trains a gigantic 
% language model on unlabeled text corpus, and then fine-tunes it on a few labeled examples at a downstream task. 
%Similar to pre-trained NLP models like BERT, computer vision pre-training is also witnessing a paradigm shift of scaling up to excessive capacity. 
% Certainly, 

The extraordinary cost of pre-training can be amortized by transferring to many downstream tasks. However, such explosive sizes of pre-trained models can even make fine-tuning computationally demanding, urging us to ask: \textit{can we aggressively trim down the complexity of pre-trained models, without damaging their downstream transferability?} Note that, the question asked is drastically different from the conventional scope of model compression \cite{han2015deep} in CV, where a model is trained, compressed and/or tuned on the same dataset and specific task. In comparison, any simplification for a pre-trained model has to ensure its intact transferability to a variety of possible downstream tasks.

To address this research gap, we turn our attention to \textit{lottery ticket hypothesis} (LTH) \cite{dettmers,evci2019rigging,frankle2018the,snip,grasp,You2020Drawing}, a fast-rising field that investigates the sparse trainable subnetworks within full dense networks. The original LTH \cite{frankle2018the,frankle2019linear} demonstrated small-scale networks contain sparse \textit{matching subnetworks} capable of training in isolation from initialization to full accuracy. In other words, we could have trained smaller networks from the start if only we had known which subnetworks to choose. Recent investigations \cite{morcos2019one,mehta2019sparse} showed those matching subnetworks to transfer between related classification tasks. However, no study has closely examined the tantalizing possibility of \textit{universal transferability} in LTH for CV models, i.e., if we treat the pre-trained weights as our initialization, \textit{whether matching subnetworks still exist in the pre-training models, that also enjoy the same downstream transfer performance? Are there universal subnetworks that can transfer to many tasks with no degradation in performance?}

The paper carries out the first comprehensive experimental study to seek these desired universal matching subnetworks, from both supervised and self-supervised pre-trained CV models. Our principled methodology bridges pre-training and LTH from two perspectives: i) \underline{Initialization via pre-training}. In the previous larger-scale settings of LTH for CV \cite{frankle2018the,Renda2020Comparing}, the matching subnetworks are found at an early point in training. Instead, we aim to identify these matching subnetworks from dense pre-trained models (self-supervised or supervised), which creates an initialization directly amenable to sparsification. ii) \underline{Transfer learning.} Finding the matching subnetwork is an expensive investment, usually costing multiple rounds of pruning and re-training. To justify this extra investment, the found subnetwork must be able to be reused by various downstream tasks, as illustrated in Figure~\ref{fig:framework}.

\begin{table*}[t]
\centering
\vspace{-1em}
\caption{Details of pre-training and fine-tuning. We use the default implementations and hyperparameters \cite{Renda2020Comparing,chen2020simple,he2020momentum,chen2020improved,chen2018encoder,lee2019drop,bochkovskiy2020yolov4}. The evaluation metrics also follow the standards \cite{Renda2020Comparing,chen2020improved,chen2018encoder,bochkovskiy2020yolov4}. For the supervised learning, we use the training dataset to name the corresponding task for the same of simplicity, e.g. ``ImageNet" represents the supervised pre-training classification task on ImageNet.}
\label{table:settings}
\begin{adjustbox}{width=1\textwidth}
\begin{threeparttable}
\begin{tabular}{l|cccccccccc}
\toprule
\multirow{2}{*}{Settings} & \multicolumn{3}{c}{Pre-training} & \multicolumn{5}{c}{Downstream Classification} & \multicolumn{1}{c}{Downstream Detection} & \multicolumn{1}{c}{Downstream Segmentation}\\ \cmidrule(lr){2-4} \cmidrule(lr){5-9} \cmidrule(lr){10-10} \cmidrule(lr){11-11} 
& ImageNet & simCLR & MoCov2 & CIFAR-10 & CIFAR-100 & SVHN & Fashion-MNIST & VisDA2017 & Pascal VOC2012/2007 & Pascal VOC 2012\\ \midrule
\# Epochs/Iters & 10 & 10 & 10 & 182 & 182 & 182 & 182 & 20 & 50 Epochs/103K Iters & 30K Iters\\ 
Batch Size & 256 & 256 & 256 & 256 & 256 & 256 & 256 & \multicolumn{1}{c}{128} & 8 & 4\\ 
\multirow{2}{*}{Learning Rate} & \multicolumn{1}{c}{0.0001} & 0.0001 & 0.0003 & 0.1 & 0.1 & 0.1 & 0.1 & \multicolumn{1}{c}{0.001} & 0.0001 & 0.01\\ 
 & \multicolumn{3}{c}{Fixed schedule} &  \multicolumn{4}{c}{$\times0.1$ at 91,136 epoch} & \multicolumn{1}{c}{$\times0.1$ at 10 epoch} &  \begin{tabular}[c]{@{}c@{}} Cosine decay \\ from $10^{-4}$ to $10^{-6}$ \end{tabular} & \begin{tabular}[c]{@{}c@{}} Linear warmup 100 Iters \\ $\times0.1$ at 18K, 22K Iters \end{tabular} \\ 
Optimizer & \multicolumn{10}{c}{SGD \cite{ruder2016overview} with 0.9 momentum}\\ 
Weight Decay & \multicolumn{1}{c}{$1\times10^{-4}$} & $1\times10^{-4}$ & $1\times10^{-4}$ & \multicolumn{1}{c}{$2\times10^{-4}$} & $2\times10^{-4}$ & $2\times10^{-4}$ & $2\times10^{-4}$ & \multicolumn{1}{c}{$5\times10^{-4}$} & $5\times10^{-4}$ & $1\times10^{-4}$\\ 
Eval. Metric & Accuracy & \multicolumn{2}{c}{Retrieval Accuracy} & Accuracy & Accuracy & Accuracy & Accuracy & Accuracy & AP, AP$_{50}$, AP$_{75}$ & mIOU\\
\bottomrule
\end{tabular}
\end{threeparttable}
\end{adjustbox}
\vspace{-1em}
\end{table*}

%The transferability is the signature attributes of pre-training \cite{chen2020simple,he2020momentum}: the extraordinary cost of pre-training is amortized by transferring to a broad range of downstream vision tasks. Locating the matching subnetworks is also an expensive investment since it usually requires multiple times repeated training for better results \cite{frankle2018the}. Fortunately, it is possible to tolerate if the resulting subnetworks are capable of reusing for diverse downstream tasks as shown in Figure~\ref{fig:framework}.

%To reduce the cost of fine-tuning on downstream tasks, we dedicate particular attention to the universal transfer behave of these subnetworks found on the supervised and self-supervised pre-training tasks. 
The course of this study presents the following findings:
\begin{itemize}
\vspace{-0.5em}
    \item Using iterative unstructured magnitude pruning \cite{frankle2018the}, we identify matching sub-networks up to $67.23\%$, $59.04\%$, $95.60\%$ sparsity, at pre-trained weights from ImageNet-equipped supervised  pre-training, simCLR and MoCo, respectively. We also find matching subnetworks at pre-trained initialization with sparsity from $73.79\%$ to $98.20\%$ in a variety of classification, detection and segmentation downstream tasks.
    \vspace{-0.3em}
    \item Subnetworks at $67.23\%$, $59.04\%$ and $59.04\%$ sparsity, found respectively using supervised ImageNet, simCLR and MoCo  pre-training, are \textit{universally} transferable to diverse downstream classification tasks with nearly the same accuracies.\vspace{-0.3em}
    \item Subnetworks at $73.79\%$/$48.80\%$, $48.80\%$/$36.00\%$ and $73.79\%$/$83.22\%$ sparsity, found respectively by supervised ImageNet, simCLR and MoCo, can transfer to downstream detection/segmentation tasks without sacrificing performance. 
    % Unfortunately, transferring these subnetworks to downstream detection tasks will incur performance degradation.  
    \vspace{-0.3em}
    \item Unlike previous matching subnetworks found at random initialization or early in training, we show that those identified at pre-trained initialization are more sensitive to structure perturbations. Also, different pre-training ways tend to yield diverse mask structures and perturbation sensitivities. 
    \vspace{-0.3em}
    \item Lastly, pruning from larger pre-trained models can also produce better transferable matching subnetworks.
    \vspace{-0.5em}
\end{itemize}
Practically speaking, this work sets the first step toward replacing large pre-trained models with smaller subnetworks, enabling much more efficient downstream tuning without inhibiting transfer performance. 
%\TL{TL: Shall we delete the following sentence?} However, \textbf{the problem is far from being fully resolved}, and our comparisons on classification/segmentation versus detection tasks reveal more subtlety that invites follow-up studies. 
As pre-training becomes increasingly central in the CV field, our results shed light on the relevance of LTH in this new paradigm. 

\vspace{-0.2em}
\section{Related Works}
\vspace{-0.5em}
\paragraph{Pruning and Lottery Tickets Hypothesis.} A trained deep network could be pruned of excess capacity \cite{lecun1990optimal}. Pruning algorithms can be grouped into unstructured  \cite{magnitude,lecun1990optimal,han2015deep} and structured 
\cite{liu2017learning,he2017channel,zhou2016less}: the former sparsifies based on weight magnitudes; while the latter considers hardware-friendliness by removing channels and so on.

The discovery of LTH \cite{frankle2018the} deviates from the convention of after-training pruning, and points to the existence of independently trainable sparse subnetworks from scratch that can match the performance of dense networks. Follow-up investigations \cite{liu2018rethinking, gale2019state} scale up LTH by rewinding approaches \cite{frankle2019lottery,Renda2020Comparing}, that re-initializes the subnetwork from the early training stage checkpoint rather than from scratch. LTH has been widely explored in image classification \cite{frankle2018the,liu2018rethinking,grasp,evci2019difficulty,Frankle2020The,savarese2020winning,yin2020the,You2020Drawing,ma2021good,chen2021long}, natural language processing \cite{gale2019state,yu2019playing,Renda2020Comparing,prasanna2020bert,chen2020lottery}, generative adversarial networks \cite{chen2021gans,chen2021ultra}, graph neural networks \cite{chen2021unified}, and reinforcement learning \cite{yu2019playing}. Most of them adopt (iterative) unstructured weight magnitude pruning \cite{han2015deep,frankle2018the}. \cite{mehta2019sparse,morcos2019one,desai2019evaluating} pioneer to study the transferability of the subnetworks identified on one image classification task to another. However, studying the universal transferability of LTH at pre-trained initializations among diverse CV tasks remains untouched. 

%Our work for the first time explores LTH in the paradigm of pre-training and fine-tuning in vision, and comprehensive downstream tasks, i.e., classification, detection and segmentation, are considered. 

One most relevant work \cite{chen2020lottery} to ours is from the natural language processing (NLP) field: the authors found universally transferable sparse matching subnetworks (at 40\% to 90\% sparsity), from the pre-trained initialization of BERT models \cite{devlin2018bert}. Finding their work inspiring, we stress that transplanting their NLP findings to our CV models is \textbf{highly nontrivial} due to multiple barriers: (1) pre-training BERT in \cite{chen2020lottery} uses only a self-supervised objective called ``masked language model" (MLM) \cite{devlin2018bert}, while pre-training CV models has a significant variety of popular options, ranging from the supervised fashion \cite{huh2016makes}, to self-supervision yet with numerous objectives \cite{doersch2015unsupervised,he2020momentum,chen2020simple}; (2) BERT models consist of self-attention and fully-connected sub-layers, differing much from the standard convolutional architectures in CV; (3) further complicating the issue is that different CV downstream tasks are known to rely on different priors and invariances; for example, while classification often calls on shift invariance, detection assumes location shift equivariance \cite{worrall2017harmonic,manfredi2020shift}. That questions the feasibility of asking for one mask to transfer among them all. Such complicacy is well manifested by our delicate observations.

%Our work is fundamentally different in that we seek universal subnetworks from supervised and self-supervised pre-training in computer vision, which can transfer to diverse downstream tasks without sacrificing performance, including classification, detection and segmentation, 

\vspace{-1em}
\paragraph{Pre-training in Computer Vision.} Supervised ImageNet pre-training has been a main CV workhorse \cite{girshick2014rich,he2019rethinking}. The recent surge of self-supervised pre-training suggest the potential of unlabeled data; examples include recovering the artificially corrupted inputs  \cite{pathak2016context,vincent2008extracting,zhang2016colorful,zhang2017split}, predicting pseudo-labels \cite{doersch2015unsupervised,dosovitskiy2014discriminative,noroozi2016unsupervised,pathak2017learning,caron2018deep,caron2019unsupervised,chen2020adversarial}, or contrasting augmented views \cite{bachman2019learning,henaff2019data,hjelm2018learning,oord2018representation,tian2019contrastive,wu2018unsupervised,zhuang2019local,chen2020simple,he2020momentum,chen2020big,chen2020improved,jiang2020robust,you2020graph}. The state-of-the-art simCLR \cite{chen2020simple,chen2020big} and MoCo \cite{he2020momentum,chen2020improved} pre-training can reduce the amount of labels needed  for tuning downstream image classifiers, by two magnitudes.

%Recently, several studies \cite{bachman2019learning,henaff2019data,hjelm2018learning,oord2018representation,tian2019contrastive,wu2018unsupervised,zhuang2019local,chen2020simple,he2020momentum,chen2020big,chen2020improved} present superior results on self-supervised pre-training using methods related to the contrastive learning \cite{hadsell2006dimensionality} which learns representations by contrasting positive pairs against negative pairs. Among this big party, two representative approaches, simCLR \cite{chen2020simple,chen2020big} and MoCo \cite{he2020momentum,chen2020improved}, show the most promising performance, which are adopted in our paper.

Pre-trained networks are usually subsequently fine-tuned, with the architectures unchanged. One exception is \cite{cai2020tiny} which is the first to adapt the backbone architecture to fit different target datasets. It pre-trains a large super-net that contains many weight-shared sub-nets that can individually operate.

\vspace{-0.5em}
\section{Preliminaries and Setups} 
\vspace{-0.5em}
In this section, we provide the detailed experimental settings and our approaches to find matching subnetworks.

\vspace{-1.2em}
\paragraph{Network.} 
We use the official ResNet-50 \cite{he2016deep} network architecture as our default backbone, while we will later compare on ResNet-152 in Section 5.2. For a particular classification downstream task, a task-specific final linear layer is added following \cite{chen2020simple}. YOLOv4 \cite{bochkovskiy2020yolov4} and DeepLabv3+ \cite{chen2018encoder} are adopted for the detection and segmentation downstream tasks respectively, which also take ResNet-50 as the backbone\footnote{For complicated CV tasks, the large variety of model design options may possibly impact our observation. For example. object detectors fall under two-stage and one-stage categories, the former often achieving higher accuracy while the latter typically being faster. YOLOv4 \cite{bochkovskiy2020yolov4} is a popular one-stage detector. We also include the results for two popular two-stage detectors, Faster RCNN \cite{ren2015faster} and SSD \cite{liu2016ssd}, in Section~\ref{sec:more_detectors}.}. Due to the various input and output scales, the first convolution layer in ResNet-50 and all classification, detection, segmentation heads are never pruned. Specifically, we let $f(x;\theta,\gamma)$ be the output of a ResNet-50 model with parameters $\theta\in\mathbb{R}^{d_1}$ (excluding the first convolution layer) and task-specific parameters $\gamma\in \mathbb{R}^{d_2}$ on an input image $x$. 

%\TL{TL: Shall we delete the following paragraph?} Note that, for complicated computer vision tasks such as object detection, the large variety of model design options may likely have impact on our observation. For example, object detection models are well-known to fall under two-stage and one-stage categories, the former often achieving higher accuracy while the latter typically being much faster. Our current study prioritizes the two-stage object detectors, using Faster RCNN as a subject example. However, we plan to expand our experiments to \textbf{one-stage detectors} soon, and plan to report new results in the short foreseeable future.

\vspace{-1.2em}
\paragraph{Pre-training.} 
For the supervised pre-training, we use the official pre-trained ResNet-50\footnote{The official Pytorch model zoo at \url{https://pytorch.org/docs/stable/torchvision/models.html}} on the ImageNet dataset \cite{deng2009imagenet}. For the self-supervised, we adopt the pre-trained ResNet-50 models with simCLR\footnote{The official simCLR model zoo at \url{https://github.com/google-research/simclr}} \cite{chen2020simple} and MoCov2\footnote{The official MoCov2 model zoo at \url{https://github.com/facebookresearch/moco}} \cite{chen2020improved} on ImageNet. 

\vspace{-1.2em}
\paragraph{Datasets, Training and Evaluation.} 
All pre-training experiments are conducted on ImageNet. For downstream tasks, we consider classification, object detection and semantic segmentation on multiple datasets. We use four natural image and one synthetic datasets to verify the transferability on classification: Fashion-MNIST \cite{xiao2017fashion}, SVHN \cite{netzer2011reading}, CIFAR-10 \cite{krizhevsky2009learning}, CIFAR-100 \cite{krizhevsky2009learning}, and VisDA2017 \cite{peng2017visda}. These datasets vary remarkably in terms of sample size, color space, resolution, image source, and classes. Following \cite{he2020momentum,chen2020improved}, we train object detection models on the combined training and validation set of Pascal VOC 2012 \cite{everingham2015pascal} and Pascal VOC 2007 \cite{everingham2010pascal}, then evaluate them on the Pascal VOC 2007 test set. We train and evaluate semantic segmentation models on Pascal VOC 2012 training and validation sets. We follow the standard hyperparameters and evaluation metrics\footnote{For detection experiments, we report the other evaluation metrics, AP$_{50}$ and AP$_{75}$ \cite{chen2020improved} in the supplement. The technical details of calculating the retrieval accuracy for simCLR and MoCo pre-training tasks are also included in the supplement.} for all pre-training and downstream tasks, as in Table~\ref{table:settings}.

\vspace{-1.2em}
\paragraph{Subnetworks.} 
% \SL{[notation issue: confused about $f(x;\theta,\cdot)$, $f(x;\theta,\gamma)$ and $f(x;\theta)$.]}
For a network $f(x;\theta,\cdot)$ with task-specific modules $\gamma$, its subnetworks can be depicted as $f(x;m\odot\theta,\cdot)$ with a pruning binary mask $m\in\{0,1\}^{d_1}$, where $\odot$ is the element-wise product. Let $\mathcal{A}^{\mathcal{T}}_t(f(x;\theta,\gamma))$ be a training algorithm (e.g., SGD with certain hyperparameters) that trains a network $f(x;\theta,\gamma)$ on a task $\mathcal{T}$ (e.g., CIFAR-10) for $t$ iterations.
% \SL{In the following, we may omit $t$ in $\mathcal{A}^{\mathcal{T}}_t$ for ease of notation.}
Let $\theta_p\in\{\theta_{\mathrm{Img}},\theta_{\mathrm{sim}},\theta_{\mathrm{MoCo}}\}$ be the pre-trained weights on ImageNet, where $\theta_{\mathrm{Img}}$ is the supervised pre-trained weight, $\theta_{\mathrm{sim}}$ and $\theta_{\mathrm{MoCo}}$ are from the self-supervised pre-training by simCLR \cite{chen2020simple} and MoCov2 \cite{chen2020improved}. Let $\theta_0$ be the random initialization, and $\theta_i$ be the network weights at the $i_{\mathrm{th}}$ epoch which is trained from $\theta_0$. Let $\mathcal{E}^{\mathcal{T}}(f(x;\theta,\gamma))$ be the evaluation function of model $f$ returned from $\mathcal{A}^{\mathcal{T}}_t$ on the corresponding task $\mathcal{T}$. Below we define:

\textit{1. Matching subnetworks.} 
Following the definition in \cite{frankle2019linear,chen2020lottery}, a subnetwork $f(x;m\odot\theta,\gamma)$ is \textit{matching} if it satisfies the following condition:
\begin{align}
   \hspace*{-0.15in} 
   %\begin{array}{l}
        \mathcal{E}^\mathcal{T}\left(\mathcal{A}^\mathcal{T}_t\left(f\left(x; m \odot \theta, \gamma\right)\right)\right) \geq \mathcal{E}^\mathcal{T}\left(\mathcal{A}^\mathcal{T}_t\left(f\left(x; \theta_p, \gamma\right)\right)\right)
          \hspace*{-0.15in}
          \label{eq:matching}
       % \vspace{-1em}
  %  \end{array} 
\end{align}
That is, matching subnetworks \underline{perform no worse than} the full dense models under the same training algorithm $\mathcal{A}_t^{\mathcal{T}}$ and evaluation metric $\mathcal{E}^{\mathcal{T}}$.

\textit{2. Winning ticket.} If $f(x;m\odot\theta,\gamma)$ is a matching subnetwork with $\theta=\theta_p$ for $\mathcal{A}_t^{\mathcal{T}}$, it is a \textit{winning ticket} for $\mathcal{A}_t^{\mathcal{T}}$.

\begin{figure}[t] 
\centering
\vspace{-0.5em}
\includegraphics[width=0.9\linewidth]{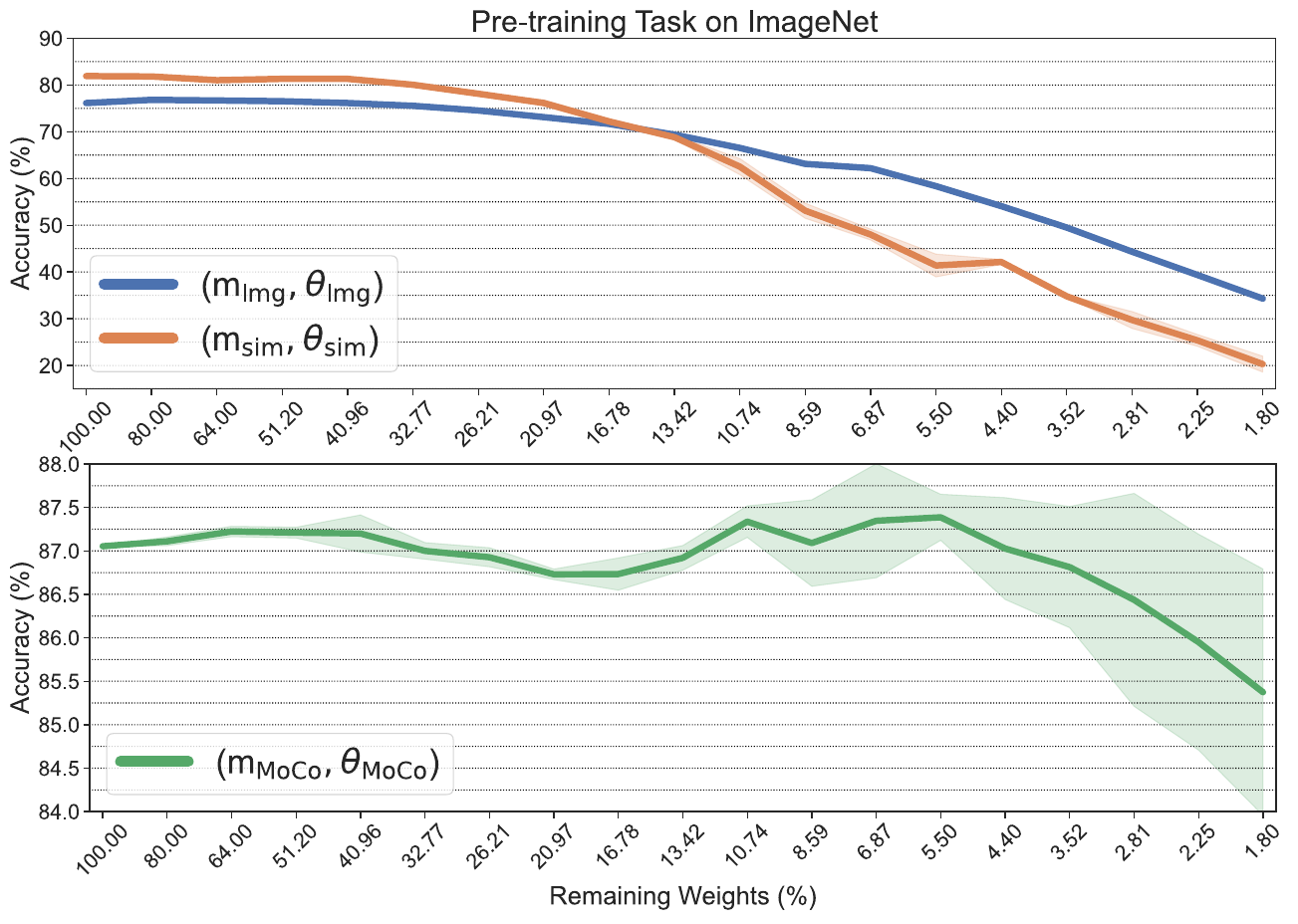}
\vspace{-0.5em}
\caption{Performance of pre-training tasks on ImageNet. The masks ($m_{\mathrm{Img}}$, $m_{\mathrm{sim}}$ and $m_{\mathrm{MoCo}}$) of evaluated subnetworks are found on supervised, simCLR and MoCo pre-training tasks respectively by IMP. $\theta_{\mathrm{Img}}$ = the pre-trained weights from the supervised ImageNet classification; $\theta_{\mathrm{sim}}$ = the pre-trained weights of simCLR \cite{chen2020simple}; $\theta_{\mathrm{MoCo}}$ = the pre-trained weights of MoCo \cite{chen2020improved}.}
\vspace{-1em}
\label{fig:pre-training}
\end{figure}

\textit{3. Universal subnetwork.} A subnetwork $f(x; m \odot \theta, \gamma_{\mathcal{T}_i})$ with task-specific configurations of $\gamma_{\mathcal{T}_i}$, is \emph{universal} for tasks $\{\mathcal{T}_i\}_{i=1}^{N}$ if and only if it is matching for each $\mathcal{A}^{\mathcal{T}_i}_{t_i}$. The task set $\{\mathcal{T}_i\}_{i=1}^{N}$ could be a group of (diverse) downstream tasks, such as classification, detection and segmentation.

\begin{figure*}[t] 
\centering
\vspace{-1em}
\includegraphics[width=0.99\linewidth]{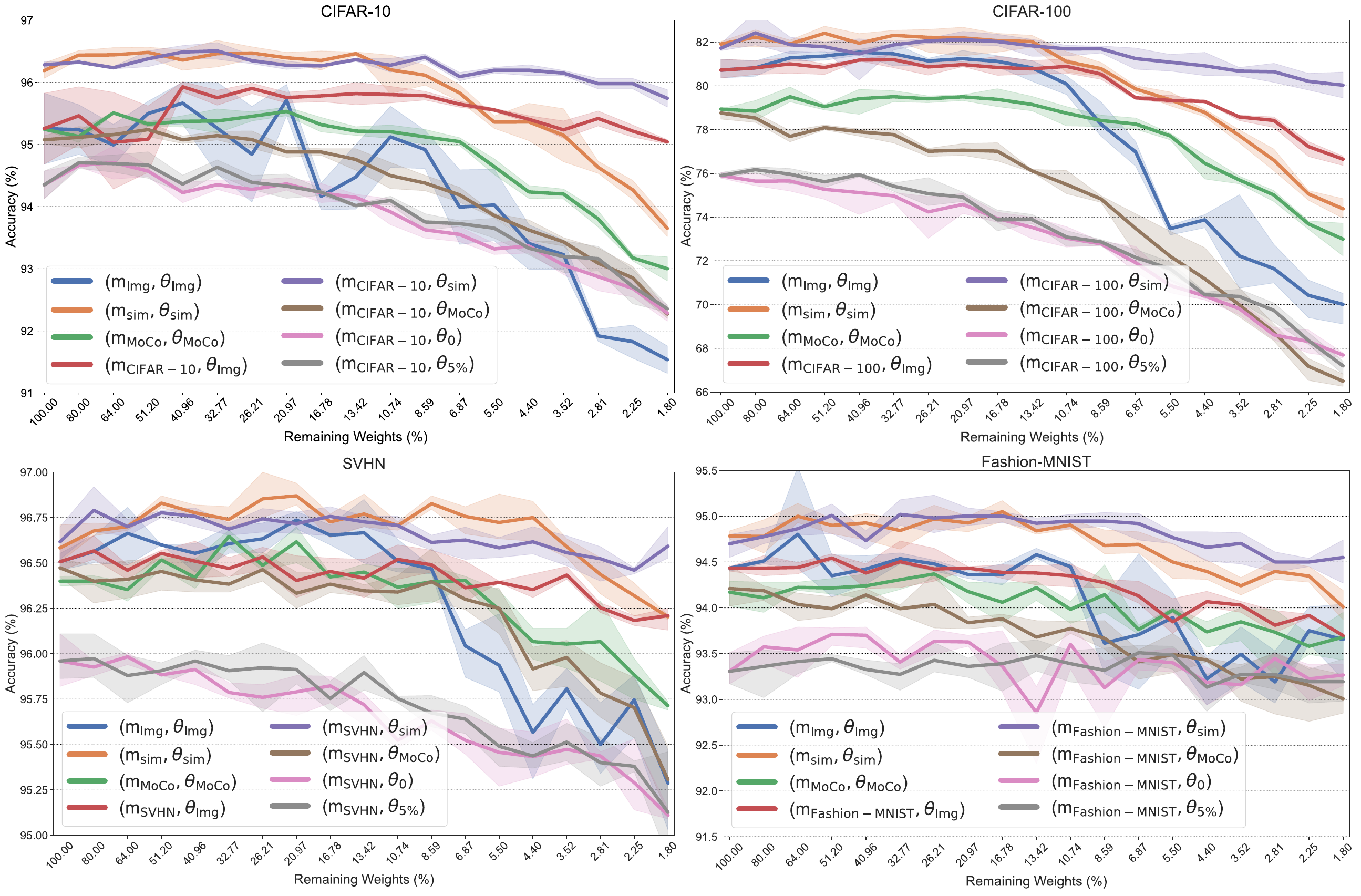}
%\vspace{-0.5em}
\caption{Performance of IMP subnetworks with a range of sparsity from $0.00\%$ to $98.20\%$ (i.e., remaining weight from $100\%$ to $1.80\%$) on downstream classification tasks, including CIFAR-10, CIFAR-100, SVHN and Fashion-MNIST. ($m_{\mathrm{Img}},\theta_{\mathrm{Img}}$), ($m_{\mathrm{sim}},\theta_{\mathrm{sim}}$) and ($m_{\mathrm{MoCo}},\theta_{\mathrm{MoCo}}$) denote transfer performance of subnetworks found at pre-training tasks. Subnetworks with ($m_{\mathcal{T}_i},\theta_{p}$), $\mathcal{T}_i\in$\{CIFAR-10, CIFAR-100, SVHN, Fashion-MNIST\} and $\theta_p\in\{\theta_{\mathrm{Img}},\theta_{\mathrm{sim}},\theta_{\mathrm{MoCo}}\}$ are identified on the downstream task $\mathcal{T}_i$ with pre-trained weights $\theta_p$. Subnetworks ($m_{\mathcal{T}_i},\theta_{0}$) and ($m_{\mathcal{T}_i},\theta_{5\%}$) are found on the task $\mathcal{T}_i$ with the random initialization $\theta_0$ \cite{frankle2018the} and an early rewinding weights $\theta_{5\%}$ \cite{Renda2020Comparing}. Curves with errors (shadow regions) are the average across three independent runs, with the standard deviations: same hereinafter.}
\vspace{-0.5em}
\label{fig:classification}
\end{figure*}

\vspace{-1em}
\paragraph{Pruning Methods.} To find the subnetworks $f(x; m \odot \theta, \gamma)$, we adopt the classical iterative magnitude pruning (IMP) approach that is commonly used by the LTH literature \cite{frankle2018the,frankle2019linear,chen2020lottery}. We prune the network by first training the unpruned dense network to completion on a task $\mathcal{T}$ (i.e., applying $\mathcal{A}^{\mathcal{T}}_t$) and then removing a portion of weights with the globally smallest magnitudes \cite{han2015deep,Renda2020Comparing}. As revealed by previous works, in order to identify the most competitive matching subnetworks, the process needs to be iteratively repeated for several rounds. Algorithm~\ref{alg:imp} outlines the full IMP procedure in the supplement.

Although beyond the current scope, our future work plans to examine the practical speedup results on a hardware platform for our training and/or inference phases. For example, in the range of 70\%-90\% unstructured sparsity, XNNPACK \cite{elsen2020fast} has already shown significant speedups over dense baselines on smartphone processors. Integrating structured pruning will be another future direction of our interest \cite{You2020Drawing}.

\section{Transfer of Pre-training Winning Tickets} \label{sec:transfer}
\vspace{-0.2em}
In this section, we first show that there exist winning tickets using the pre-trained initialization on both self-supervised and supervised pre-training tasks. As shown in Figure~\ref{fig:pre-training}, we find winning tickets with $67.23\%$, $59.04\%$ and $95.60\%$ sparsity for supervised ImageNet, self-supervised simCLR and MoCo pre-training tasks.

Then, we investigate to what extent IMP subnetworks found for pre-training tasks can (universally) transfer to different downstream tasks. 
%Evidenced by Chen et al. \cite{chen2020lottery}, universally transferable subnetworks can be identified in BERT pre-training. To study this possibility in CV, 
We ask the following questions: %\SL{[Ok. It seems that you use $f(x;\theta,\cdot)$ to represent the  model used for pre-training (w/o downstream task specification), and use $f(x;\theta,\gamma)$ for task-specific model built over the pre-training backbone $\theta$. Right? If so, you should define this notation in paragraph 'Network' when you first introduce  $f$ in Sec. 3. ]}

\textit{\textbf{Q1:}} Are winning tickets $f(x;m_{\mathcal{P}}\odot\theta_p,\cdot)$, found on the pre-training task $\mathcal{P}$, also winning tickets for other downstream tasks $\mathcal{T}$?

\textit{\textbf{Q2:}} Are there common patterns in the transferability of winning tickets from different pre-trainings (e.g., supervised versus self-supervised)?

\textit{\textbf{Q3:}} Can the transferred subnetworks $f(x;m_{\mathcal{P}}\odot\theta_p,\cdot)$ outperform the subnetworks $f(x;m_{\mathcal{T}}\odot\theta_i,\cdot)$ ($\theta_i\in\{\theta_0,\theta_{5\%}\footnote{Early weight rewinding \cite{Renda2020Comparing,frankle2019linear} improves the quality of found matching subnetworks. As indicated by \cite{frankle2019linear}, the best rewinding points usually lie in the first $1\%\sim 5\%$ training epochs. We take $5\%$ for default comparison.},\theta_p\}$), found on a specific task $\mathcal{T}$?

% \begin{figure}[t] 
% \centering
% \vspace{-0.5em}
% \includegraphics[width=0.9\linewidth]{Figs/Pre-train.pdf}
% \vspace{-0.5em}
% \caption{Performance of pre-training tasks on ImageNet. The masks ($m_{\mathrm{Img}}$, $m_{\mathrm{sim}}$ and $m_{\mathrm{MoCo}}$) of evaluated subnetworks are found on supervised, simCLR and MoCo pre-training tasks respectively by IMP. $\theta_{\mathrm{Img}}$ = the pre-trained weights from the supervised ImageNet classification; $\theta_{\mathrm{sim}}$ = the pre-trained weights of simCLR \cite{chen2020simple}; $\theta_{\mathrm{MoCo}}$ = the pre-trained weights of MoCo \cite{chen2020improved}.}
% \vspace{-1em}
% \label{fig:pre-training}
% \end{figure}

\begin{figure}[t] 
\centering
\vspace{-1em}
\includegraphics[width=1\linewidth]{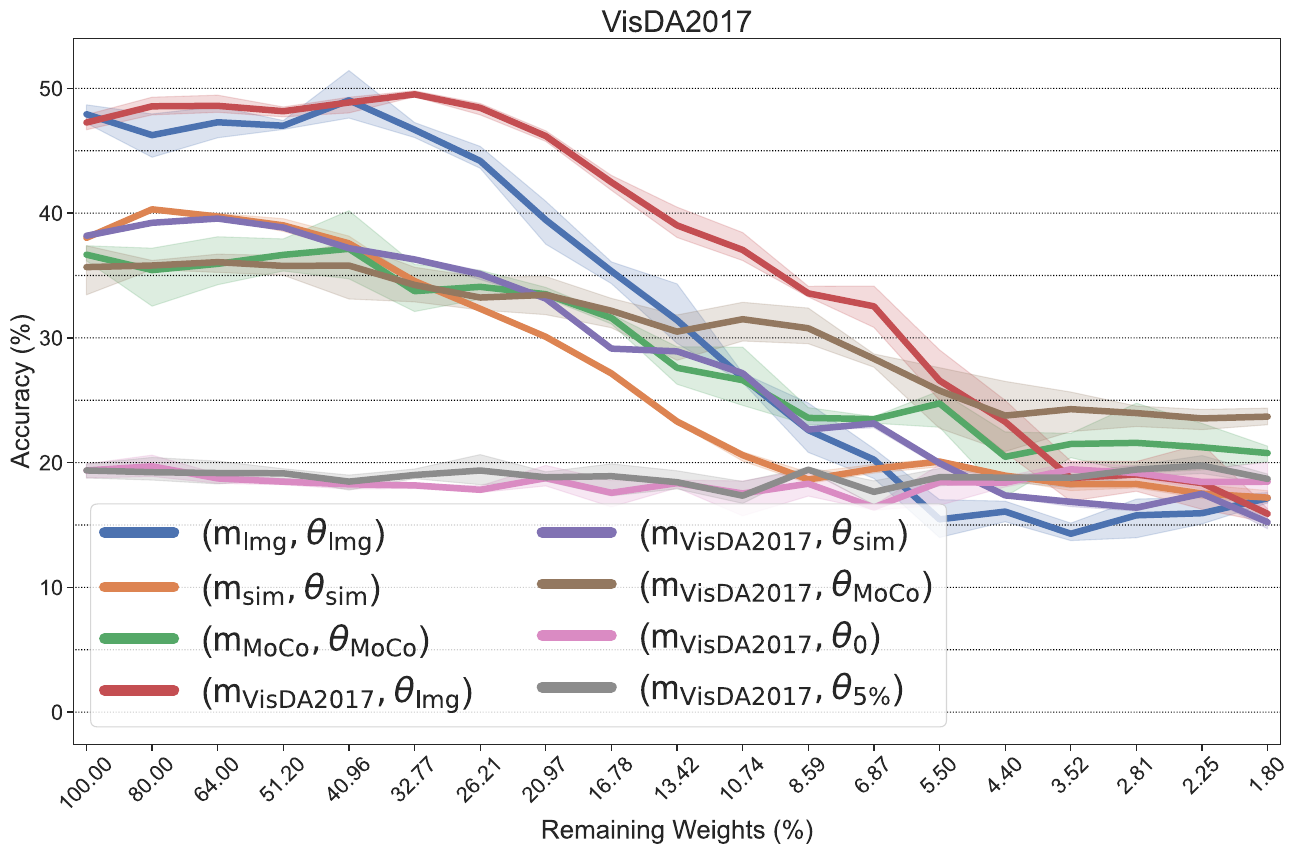}
\vspace{-1.5em}
\caption{Performance of IMP subnetworks with a range of sparsity from $0.00\%$ to $98.20\%$ on the synthetic dataset, VisDA2017.}
\label{fig:visda2017}
\vspace{-4mm}
\end{figure}

\subsection{Transfer to Classification Tasks}
As shown in Figures~\ref{fig:classification} and \ref{fig:visda2017}, evaluated subnetworks are divided into three groups, according to sources of ($m,\theta$): i) transferred subnetworks with ($m_{\mathcal{P}},\theta_p$), $\mathcal{P}\in\{\mathrm{Img},\mathrm{sim},\mathrm{MoCo}\}$ and $\theta_p\in\{\theta_{\mathrm{Img}},\theta_{\mathrm{sim}},\theta_{\mathrm{MoCo}}\}$; ii) subnetworks found on a specific downstream tasks with pre-trained weights ($m_{\mathcal{T}},\theta_p$), $\mathcal{T}\in$\{CIFAR-10, CIFAR-100, SVHN, Fashion-MNIST, VisDA2017\}; iii) subnetworks consists of ($m_{\mathcal{T}},\theta_i$), $\theta_i\in\{\theta_0,\theta_{5\%}\}$, identified with the original random initialization $\theta_0$ or early rewinding weights $\theta_{5\%}$ on downstream tasks $\mathcal{T}$. Summarizing all comprehensive results, our main observations are:

\vspace{-1em}
\paragraph{A1: Subnetworks with ($m_{\mathcal{P}},\theta_p$) universally transfer to diverse downstream classification tasks.}  As shown in Figure~\ref{fig:classification} and Figure~\ref{fig:visda2017}, compared with unpruned dense models, subnetworks found on pre-training tasks ($f(x;m_\mathrm{Img}\odot\theta_\mathrm{Img},\cdot)$, $f(x;m_\mathrm{sim}\odot\theta_\mathrm{sim},\cdot)$, $f(x;m_\mathrm{MoCo}\odot\theta_\mathrm{MoCo},\cdot)$) transfer without sacrificing performance\footnote{Practically, to account for random fluctuations, we consider a subnetwork to be a winning ticket as long as its performance is within one standard deviation of the unpruned dense model.} by sparsity ($91.41\%$, $91.41\%$, $91.41\%$) to CIFAR-10, ($86.58\%$, $86.58\%$, $89.26\%$) to CIFAR-100, ($91.41\%$, $96.48\%$, $93.13\%$) to SVHN, ($89.26\%$, $89.26\%$, $91.41\%$) to Fashion-MNIST, and ($67.23\%$, $59.04\%$, $59.04\%$) to VisDA2017. Therefore, we observe that subnetworks produced by supervised ImageNet, self-supervised simCLR and MoCo pre-training tasks, universally transfer to four downstream natural image datasets with sparsity ($86.58\%$, $86.58\%$, $89.26\%$), respectively. However, it requires larger network capacity, i.e., ($67.23\%$, $59.04\%$, $59.04\%$), to transfer to the synthetic VisDA2017 dataset without loss of performance. 

\begin{figure*}[t] 
\centering
\vspace{-0.5em}
\includegraphics[width=0.95\linewidth]{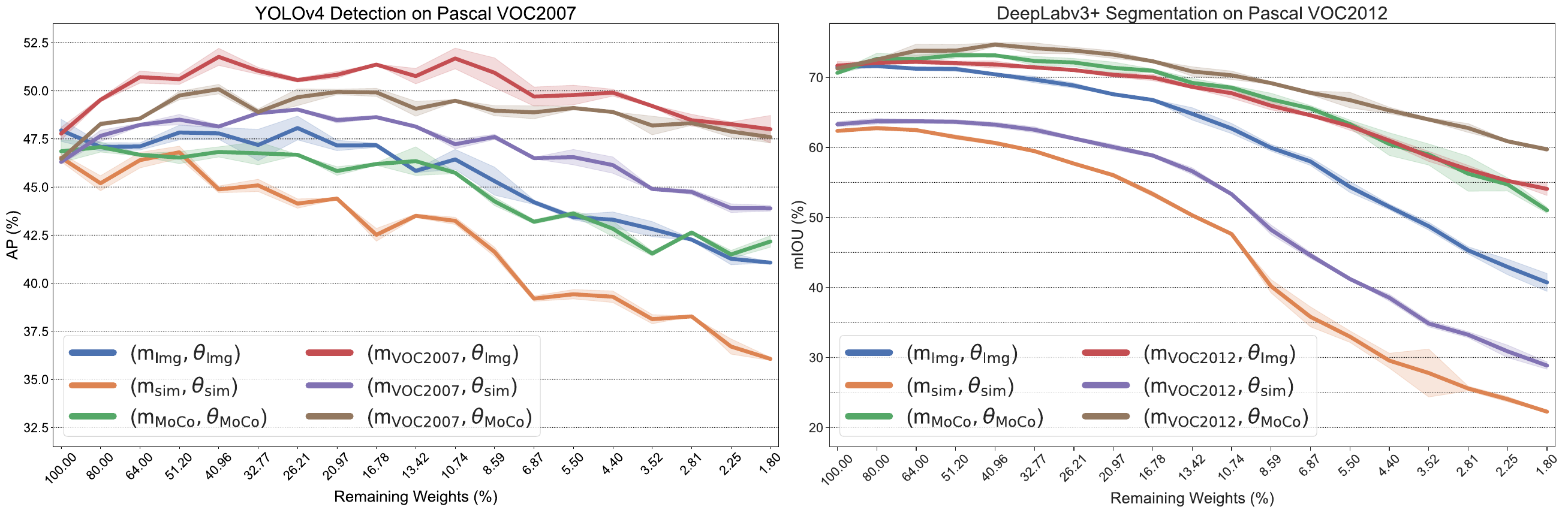}
\vspace{-0.5em}
\caption{Performance of IMP subnetworks with a range of sparsity from $0.00\%$ to $98.20\%$ on the downstream detection and segmentation tasks. Subnetworks with ($m_{\mathrm{VOC2007}},\theta_{p}$) and ($m_{\mathrm{VOC2012}},\theta_{p}$), $\theta_p\in\{\theta_{\mathrm{Img}},\theta_{\mathrm{sim}},\theta_{\mathrm{MoCo}}\}$ are identified on the downstream detection and segmentation tasks with pre-trained weights $\theta_p$, respectively. The standard deviations are around $0.1\%\sim2.5\%$ AP/mIOU.}
\vspace{-1em}
\label{fig:dense}
\end{figure*}

\vspace{-1em}
\paragraph{A2: Winning tickets from different pre-training ways, have diverse behaviors, that are also affected by the downstream task properties.} On natural image datasets, subnetworks found with self-supervised pre-training (i.e., simCLR and MoCo) outperform subnetworks found with supervised ImageNet pre-training at the extreme sparsity level (e.g., more than $93.13\%$). Specifically, $f(x;m_{\mathrm{sim}}\odot\theta_{\mathrm{sim}},\cdot)$ consistently achieves superior generalization across four downstream datasets.  $f(x;m_{\mathrm{MoCo}}\odot\theta_{\mathrm{MoCo}},\cdot)$ performs worse than $f(x;m_{\mathrm{Img}}\odot\theta_{\mathrm{Img}},\cdot)$ at the low and middle level sparsity of subnetworks. However, the conclusions are almost flipped when transferring $f(x;m_{\mathcal{P}}\odot\theta_{p},\cdot)$ to the synthetic VisDA2017 dataset. Subnetworks $f(x;m_{\mathrm{Img}}\odot\theta_{\mathrm{Img}},\cdot)$ surpass others with a large performance margin, at the sparsity from $0.00\%$ to $89.26\%$. For the extreme sparsity, the MoCo pre-training task generates a better transferable subnetworks. These observations suggest that supervised ImageNet pre-training allows subnetworks to transfer to the downstream datasets even with domain gaps to the pre-training datasets (e.g., from natural to synthetic images); self-supervised pre-trainings (e.g., simCLR and MoCo) produce more transferable subnetworks especially at the extreme sparsity, when natural image datasets are at downstream.

%which are resembling ImageNet used for pre-training tasks. 

\vspace{-1em}
\paragraph{A3: Transferred subnetworks $f(x;m_{\mathcal{P}}\odot\theta_p,\cdot)$ perform the best until extreme sparsity.} Subnetworks $f(x;m_{\mathcal{T}}\odot\theta_p,\cdot)$, found on a specific downstream task with pre-trained weights, can be considered as \textit{``performance upbound"} for all our IMP subnetworks. $f(x;m_{\mathcal{T}}\odot\theta_p,\cdot)$ is identified as matching subnetworks with the sparsity ($98.20\%$, $91.41\%$, $73.79\%$) for CIFAR-10, ($91.41\%$, $91.41\%$, $20.00\%$) for CIFAR-100, ($91.41\%$, $95.60\%$, $91.41\%$) for SVHN, ($89.26\%$, $96.48\%$, $73.79\%$) for Fashion-MNIST, and ($73.79\%$, $59.04\%$, $67.23\%$) for VisDA2007.

For universal transferable subnetworks, we observe: i) $f(x;m_{\mathrm{Img}}\odot\theta_{\mathrm{Img}},\cdot)$ and $f(x;m_{\mathrm{sim}}\odot\theta_{\mathrm{sim}},\cdot)$ match the corresponding $f(x;m_{\mathcal{T}}\odot\theta_p,\cdot)$ with at most $59.04\%$ sparsity; ii) On the natural image datasets, $f(x;m_{\mathrm{MoCo}}\odot\theta_{\mathrm{MoCo}},\cdot)$ steadily outperform $f(x;m_{\mathcal{T}}\odot\theta_p,\cdot)$ by a clear margin across all sparsity levels, especially for CIFAR-100; On the synthetic dataset, it fails to match under an excessive sparsity (i.e., $>83.22\%$). Note that subnetworks with $\theta_0$ and $\theta_{5\%}$ are inferior on all downstream tasks, compared to subnetworks with pre-trained initialization $\theta_p$.

\subsection{Transfer to Detection and Segmentation} 
\vspace{-0.5em}
Training detection and segmentation models commonly starts from pre-trained initializations  \cite{oquab2014learning,he2019rethinking,chen2020improved}. We compare the transferred subnetworks with ($m_{\mathcal{P}},\theta_p$) versus the downstream task subnetworks with ($m_{\mathcal{T}},\theta_p$), as shown in Figure~\ref{fig:dense}. Observations are organized as follows: 

% \paragraph{A1: Subnetworks $f(x;m_{\mathcal{P}}\odot\theta_p,\cdot)$ transfer to the segmentation task successfully, but NOT so on the detection task.} The winning tickets $f(x;m_{\mathcal{P}}\odot\theta_p,\cdot)$ found on the pre-training tasks are no longer matching subnetworks on the detection task, which incurs performance degradation compared to unpruned dense models $f(x;\theta_p,\cdot)$. Fortunately, we still manage to find transferable winning tickets on the segmentation task with the sparsity ($48.80\%$, $36.00\%$, $83.22\%$) for supervised ImageNet pre-training, self-supervised simCLR and MoCo pre-training tasks respectively.

\vspace{-1em}
\paragraph{A1: Subnetworks $f(x;m_{\mathcal{P}}\odot\theta_p,\cdot)$ transfer to the detection and segmentation tasks successfully.} 
% The winning tickets $f(x;m_{\mathcal{P}}\odot\theta_p,\cdot)$ found on the pre-training tasks are no longer matching subnetworks on the detection task, which incurs performance degradation compared to unpruned dense models $f(x;\theta_p,\cdot)$. Fortunately,
Figure~\ref{fig:dense} demonstrates it is manageable to find transferable winning tickets on the detection and segmentation with the sparsity ($73.79\%$, $48.80\%$, $73.79\%$) and ($48.80\%$, $36.00\%$, $83.22\%$) for supervised ImageNet pre-training, self-supervised simCLR and MoCo pre-training tasks respectively.

\vspace{-1.2em}
\paragraph{A2: Unlike classification, winning tickets from diverse pre-training tasks behave similarly on downstream detection and segmentation tasks.} In Figure~\ref{fig:dense}, we observe the evident ranking of achieved transfer performance across all sparsity levels: $\mathcal{E}^{\mathcal{T}}(f(x;m_{\mathrm{MoCo}}\odot\theta_{\mathrm{MoCo}},\cdot)) > \mathcal{E}^{\mathcal{T}}(f(x;m_{\mathrm{Img}}\odot\theta_{\mathrm{Img}},\cdot)) > \mathcal{E}^{\mathcal{T}}(f(x;m_{\mathrm{sim}}\odot\theta_{\mathrm{sim}},\cdot))$, $\mathcal{T}\in\{\mathrm{detection}, \mathrm{segmentation}\}$. It suggests that MoCo pre-trained weights are most favorable for transferring to detection and segmentation tasks \cite{chen2020improved}.

\vspace{-1.2em}
\paragraph{A3: Subnetworks $f(x;m_{\mathcal{T}}\odot\theta_p,\cdot)$ surpass subnetworks $f(x;m_{\mathcal{P}}\odot\theta_p,\cdot)$ by a non-negligible margin.} As shown in Figure~\ref{fig:dense}, with the assistance from the pre-trained initialization ($\theta_{\mathrm{Img}}$, $\theta_{\mathrm{sim}}$, $\theta_{\mathrm{MoCo}}$), we find winning tickets with the sparsity at level ($95.60\%$, $93.13\%$, $97.75\%$) and ($73.79\%$, $67.23\%$, $86.58\%$) for detection and segmentation respectively. These identified winning tickets consistently outperform transferred subnetwork with ($m_{\mathcal{P}},\theta_p$).

\begin{figure}[!ht] 
\centering
% \vspace{-0.5em}
\includegraphics[width=0.95\linewidth]{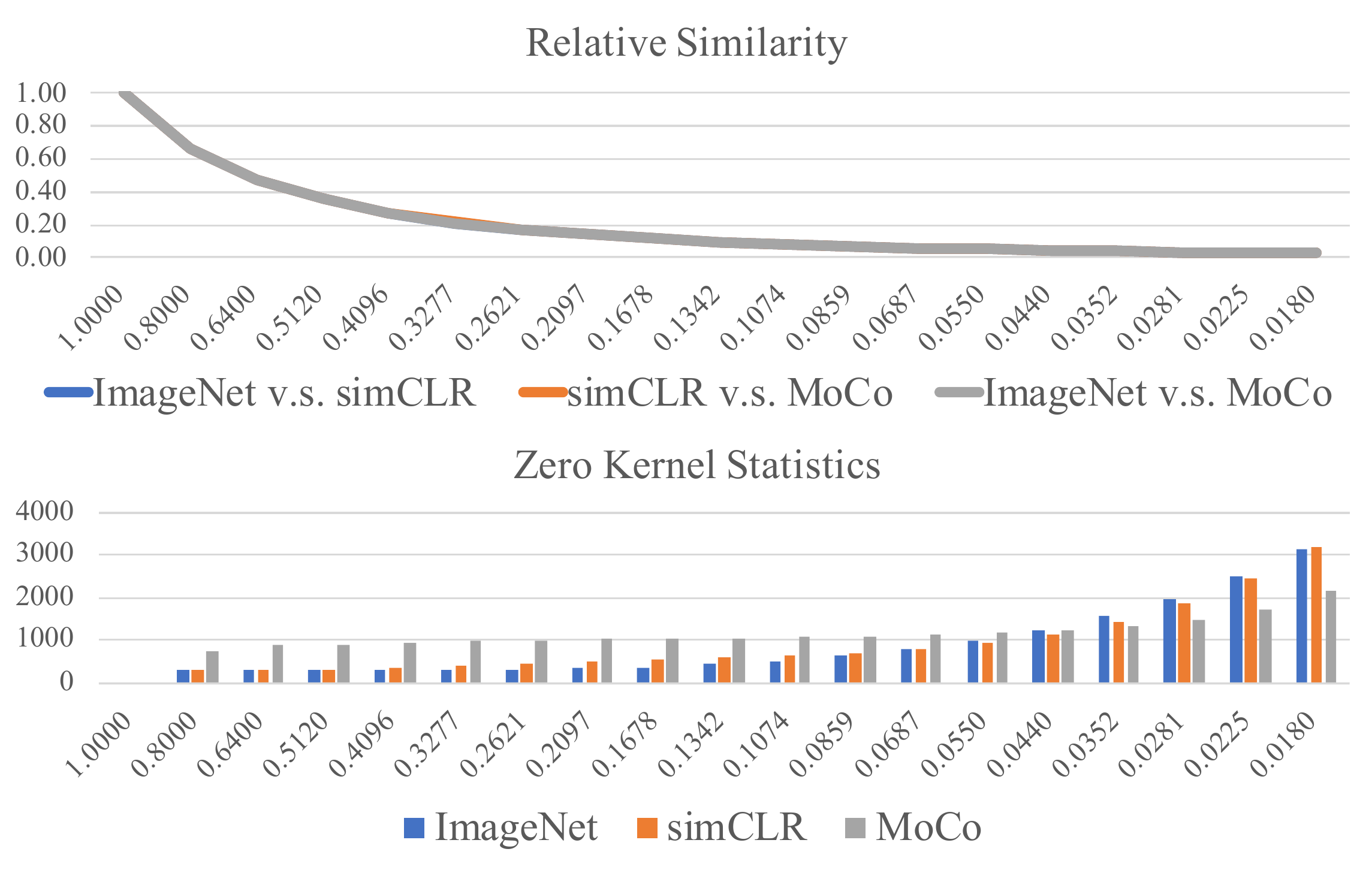}
\vspace{-0.5em}
\caption{Top: The relative mask similarity between subnetworks which identified on supervised ImageNet, simCLR and MoCo pre-training tasks. Bottom: The number of completely pruned (zero) kernels in subnetworks found on different pre-training tasks.}
\vspace{-0.5em}
\label{fig:mask}
\end{figure}

\section{Analyzing Properties of Pre-training Tickets}

\subsection{Comparing Masks from Different Pre-trainings}
In Figure~\ref{fig:mask}, we compare the overlap in sparsity patterns found for different pre-training tasks. Relative similarity (i.e., $\frac{m_i\cap m_j}{m_i \cup m_j}$ in \cite{chen2020lottery}) are reported, which reflects the overlap degree between hamming masks $m_i$ and $m_j$, where $i,j \in\{\mathrm{Img},\mathrm{sim},\mathrm{MoCo}\}$. We find that subnetworks for pre-training tasks are remarkably heterogeneous: they share less than $6.55\%$ locations in common after five-round IMP; the more sparsified, the larger differences.

We also calculate the number of completely pruned (zero) kernels of subnetworks in Figure~\ref{fig:mask}, which roughly reveals the weight clustering status in the sparse models. We observe that the remaining weights of subnetworks identified on the MoCo pre-training task are more clustered (i.e. more zero kernels) than the ones from ImageNet and simCLR, until reaching an extreme sparsity like $95.60\%$. 

\begin{figure}[t] 
\centering
\includegraphics[width=0.99\linewidth]{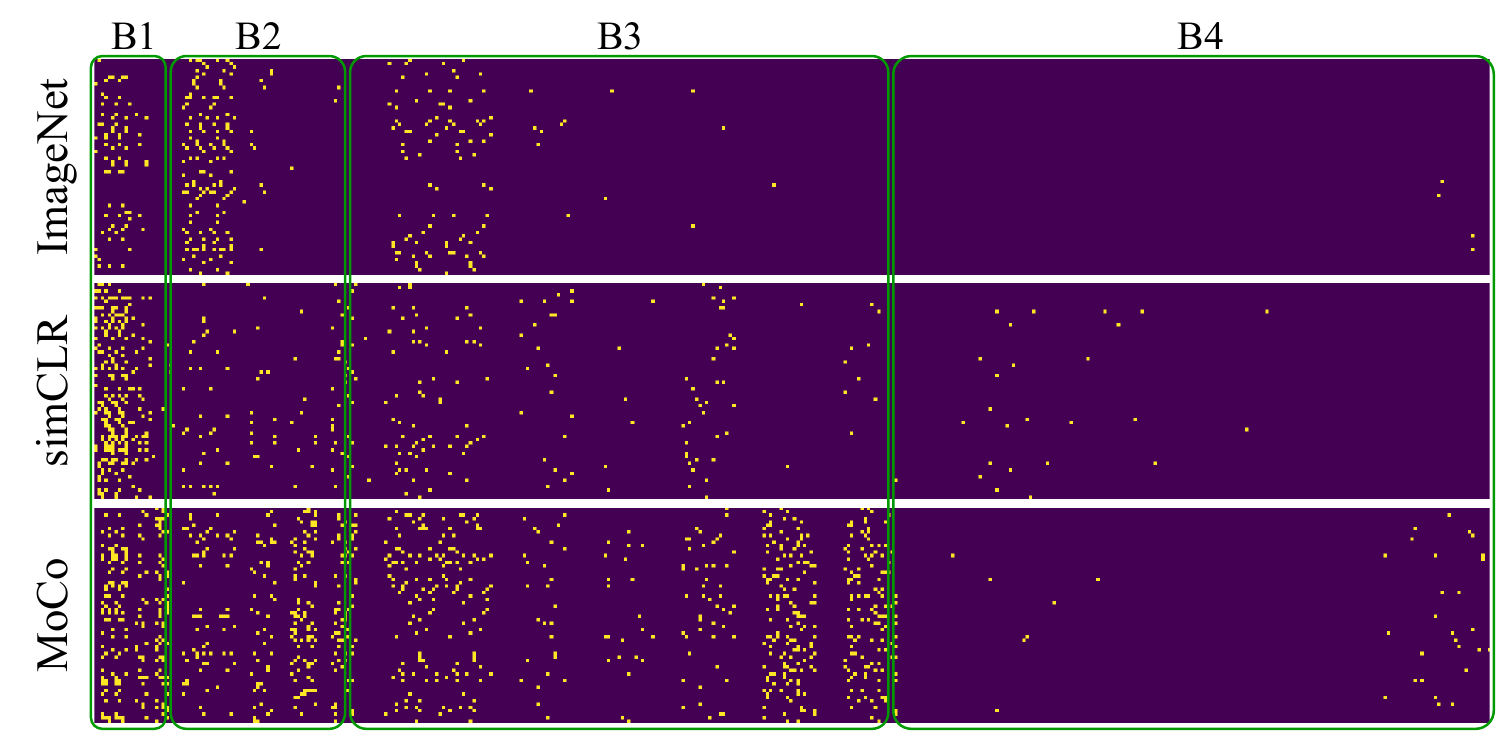}
\vspace{-0.2em}
\caption{Kernel-wise heatmap visualizations of subnetworks with $79.03\%$ sparsity found on supervised ImageNet, simCLR and MoCo pre-training tasks. From left to right, we visualization all kernels of subnetworks from the input to the output layers. The bright dots (\textcolor{yellow}{$\bullet$}) represent the completely pruned (zero) kernels and the dark dots (\textcolor{purple}{$\bullet$}) the kernels having at least one unpruned weight. B1$\sim$B4 donate four residual blocks in the ResNet-50 backbone.}
\vspace{-1em}
\label{fig:zerokenerl}
\end{figure}

Specifically, we provide kernel-wise heatmap visualizations of subnetworks with $79.03\%$ sparsity in Figure~\ref{fig:zerokenerl}. We find that the completely pruned (zero) kernels are mainly clustered in the early layers of subnetworks, and appear rarely in the later layers. Among three kinds of subnetworks, the one from MoCo has the most dispersed distribution of completely pruned kernels. In general, more structured sparse subnetworks (i.e., more all-zero kernels) may have a stronger potential for hardware speedup \cite{elsen2020fast}.

\subsection{Pre-training versus Random Initialization}

\begin{figure*}[t] 
\centering
\vspace{-1em}
\includegraphics[width=\linewidth]{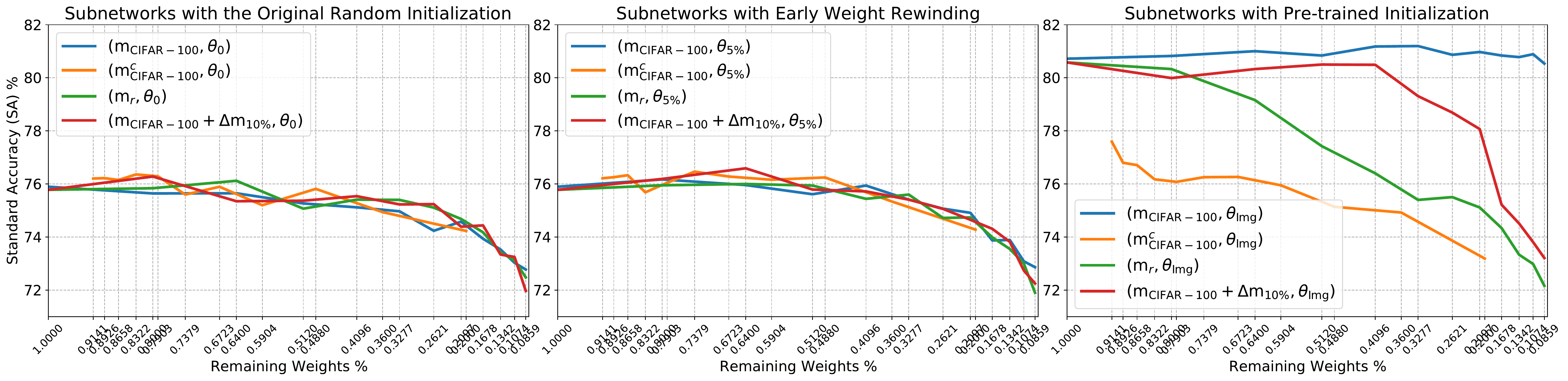}
\vspace{-1em}
\caption{Performance comparison across subnetworks found on CIFAR-100 with the original random initialization $\theta_0$, early rewinding weight $\theta_{5\%}$, and the pre-trained weights $\theta_{\mathrm{Img}}$. $m_{\mathrm{CIFAR-100}}$ = masks found by IMP; $m^c_{\mathrm{CIFAR-100}}$ = the complementary masks of $m_{\mathrm{CIFAR-100}}$, where $m\cap m^c=\emptyset$ and $m\cup m^c=\boldsymbol{1}\in\mathbb{R}^{d_1}$; $m_r$ = random pruned mask; $\Delta m_{10\%}$ = mask perturbations by randomly flipping $10\%$ ``$1$" and $10\%$ ``$0$" in the mask $m\in\{0,1\}^{d_1}$ to its opposite value. Curves are the average across three independent runs.}
\vspace{-1em}
\label{fig:ct}
\end{figure*}

%\paragraph{Identifying Subnetworks from Pre-training versus Random Initialization.} 
%\TL{Still looking for a better position to pose this analysis, it seems does not fit well here.} 
%Besides the investigation of transferable subnetworks in Section~\ref{sec:transfer}, 
A signature of our setting is to treat pre-trained weights as the initialization, in contrast to most LTH works starting from random initialization \cite{frankle2018the,frankle2019linear}. These two configurations produce matching subnetworks with diverse behaviors, including generalization performance and the structure sensitivity of obtained masks. We perform IMP on CIFAR-100 with the original random initialization $\theta_0$, early rewinding weights $\theta_{5\%}$, and the pre-trained weights $\theta_{\mathrm{Img}}$ respectively, and then generates subnetworks consisting of ($m_{\mathrm{CIFAR-100}}$, $\theta$), $\theta\in\{\theta_0,\theta_{5\%},\theta_{\mathrm{Img}}\}$. As for comparison baselines, we consider three mask variants, the complementary masks $m^c_{\mathrm{CIFAR-100}}$, randomly pruned masks $m_r$, and the perturbed masks $m_{\mathrm{CIFAR-100}}+\Delta m_{10\%}$ as in Figure~\ref{fig:ct}. Several observations can be draw as follows:
\begin{itemize}
\vspace{-0.3em}
    \item Starting from $\theta_0$ or $\theta_{5\%}$, identified subnetworks are resilient to structure perturbations. In other words, there only exist marginal performance differences across subnetworks with masks $m_{\mathrm{CIFAR-100}}$, $m^c_{\mathrm{CIFAR-100}}$, $m_r$ and $m_{\mathrm{CIFAR-100}}+\Delta m_{10\%}$. However, the found subnetworks with the pre-trained initialization behave in sharp contrast, that all complementary masks, random pruned masks and perturbed masks substantially degraded the performance w.r.t. the IMP masks. A possible explanation is that the pre-trained initializations are already highly structured, and perturbations can destroy the intrinsic structure. As evidenced by the right subfigure of Figure~\ref{fig:ct}, subnetworks with ($m^c_{\mathrm{CIFAR-100}}$,$\theta_{\mathrm{Img}}$) are no better than subnetwork with ($m_{\mathrm{CIFAR-100}},\theta_0$). It shows that pre-training with damaged weight distributions no longer leads to the generalization gains.\vspace{-0.2em}
    \item Comparing the randomly pruned subnetworks in Figure~\ref{fig:ct}, we observe that pre-trained initialization consistently benefits the accuracy until subnetworks reaching some high sparsity (e.g., $67.23\%$). After that, the performance of random pruned subnetworks is no longer affected by different initializations.
    \vspace{-0.3em}
\end{itemize}

%\paragraph{Mask Comparisons across Diverse Pre-training.} 

\subsection{More Ablation Studies for Pre-training}
\vspace*{-0.05in}
\paragraph{Larger Pre-training Model?} \cite{li2020train} reveals that heavily compressed, large transformer models achieve higher performance than lightly compressed, small transformer models in natural language processing. We re-confirm this claim for self-supervised simCLR pre-training, in terms of the transferability\footnote{In the supplement, we also report the pre-training task performance of subnetworks generated from small- and large-scale pre-trained simCLR.} of found matching subnetworks. 

\begin{figure}[!ht] 
\centering
\includegraphics[width=0.8\linewidth]{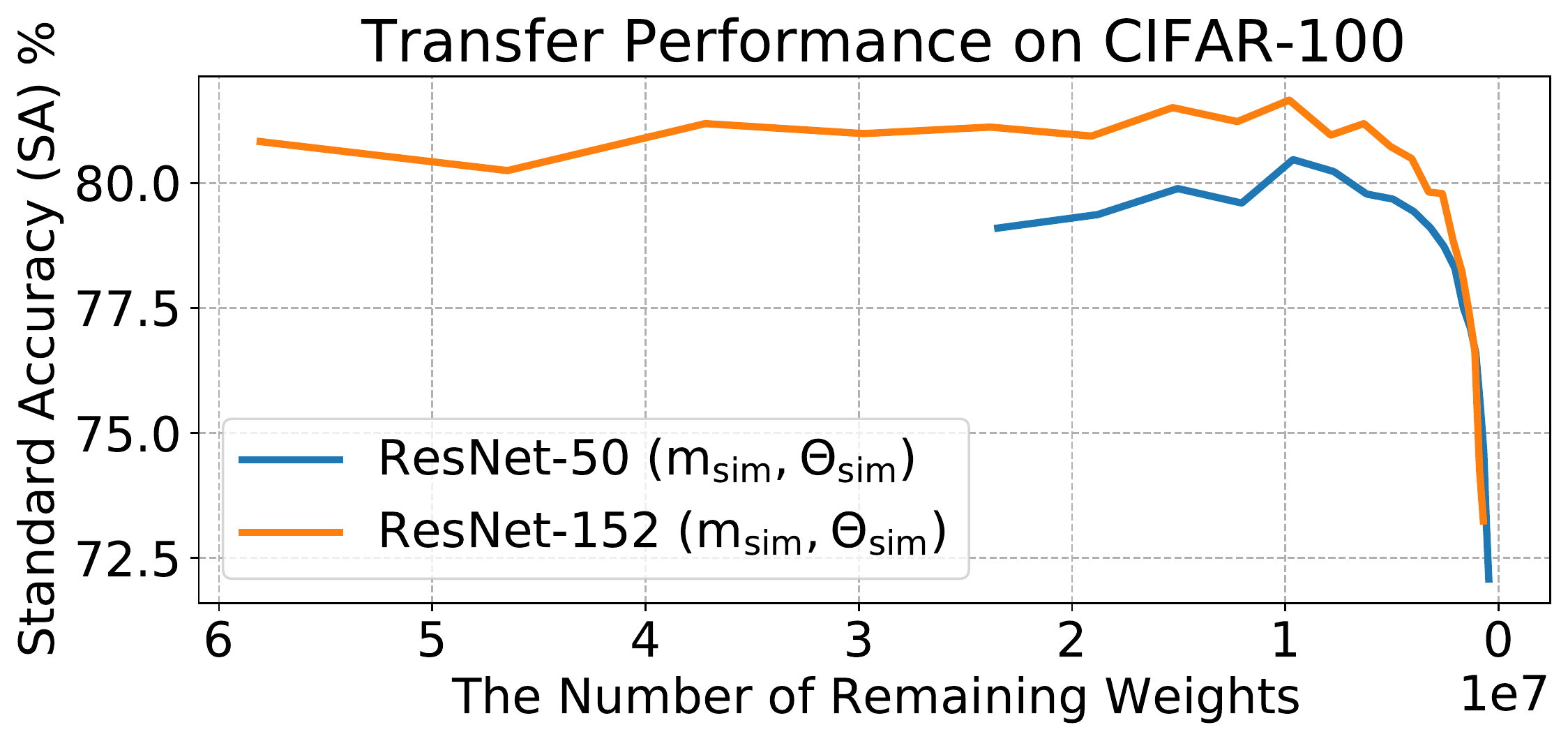}
%\vspace{-0.5em}
\caption{Transfer performance on CIFAR-100 over the number of remaining weights. Subnetworks are found on the simCLR pre-training task with pre-trained ResNet-50 and ResNet-152 weights.}
\label{fig:res152}
\vspace{-0.5em}
\end{figure}

In Figure~\ref{fig:res152}, with the same number of remaining weights, subnetworks pruned from simCLR\footnote{For a fair comparison, here we adopt the simCLRv2\cite{chen2020big} pre-trained ResNet-152 and ResNet-50 models, since only simCLRv2 released the official pre-trained ResNet-152 model.} pre-trained ResNet-152, achieve consistently superior accuracy on the downstream CIFAR-100 task than the ones from simCLR pre-trained ResNet-50 (around one-third size of ResNet-152). At least for simCLR, pruning from larger pre-trained models produces better transferable matching subnetworks. 

Our observation is also aligned with the advocates of \cite{chen2020big}, to first pretrain a big model and then compress it. The key difference is that, \cite{chen2020big} uses standard model compression (knowledge distillation) \textit{after downstream fine-tuning is done}; in contrast, our results can be seen as a possible second pre-training stage: after the initial pre-training (and before any fine-tuning), performing IMP to find equally-capable matching subnetwork with far fewer parameters. 

\vspace{-1em}
\paragraph{Temperature Hyperparameter.} The temperature scaling hyperparameter is known to play a significant role in the quality of the simCLR pre-training \cite{chen2020simple,chen2020big,chen2020intriguing}. It motivates us to investigate the impact of the temperature scaling factor on the transferability of pre-training winning tickets found in Section~\ref{sec:transfer}. Without loss of the generality, we consider the subnetworks with the sparsity from $67.23\%$ to $73.79\%$. Specifically, we start from training subnetworks at the sparsity level $67.23\%$ for $10$ epochs, on the simCLR task with different temperature scaling factors. Then, they are pruned to the level of $73.79\%$ sparsity by IMP. Finally, subnetworks are fine-tuned and evaluated on the downstream CIFAR-100 task. Results in Table~\ref{table:temp} show that found subnetworks have close transfer performance if the temperature scaling factor lies in a moderate range (i.e., [$0.1,0.5$]), and the performance will degrade at extreme temperatures (e.g., 20.0).

\begin{table}[!htb]
\vspace{-0.5em}
% \vspace*{\vspace{-0.5em}}
\centering
\caption{Ablation study of  temperature  parameter in simCLR. Transfer performance (i.e., accuracy) of subnetworks ($m_{\mathrm{sim}},\theta_{\mathrm{sim}}$) with $73.79\%$ sparsity on  CIFAR-100  downstream task.}
\label{table:temp}
\begin{adjustbox}{width=0.45\textwidth}
\begin{tabular}{l|ccccccc}
\toprule
Temperature & 0.1 & 0.2 & 0.5 & 1.0 & 2.0 & 10.0 & 20.0\\ \midrule
Accuracy (\%) & 81.81 & 81.91 & \textbf{82.22} & 81.24 & 80.76 & 81.46 & 80.18\\ \bottomrule
%\vspace{-1em}
\end{tabular}
\end{adjustbox}
\end{table}

\vspace{-1.3em}
\section{Conclusion}
\vspace{-0.5em}
We study the lottery ticket hypothesis in the context of CV pre-training, via both supervised (e.g., ImageNet classification) and self-supervised (e.g., simCLR and MoCo) ways. Despite the complicacy of our goal, by performing IMP from the pre-trained initializations, we are consistently able to find matching subnetworks at non-trivial sparsity levels, that can be independently trained to full model performance, on both pre-training and downstream tasks. We also present a detailed discussion of cross-task universal transferability. 
%Our future work plans to extend our experiments
% and observations, 
%to 
% both more model types (e.g., one-stage object detector), and 
%more computer vision tasks (e.g., 3D vision).

%Moreover, found universal subnetworks can transfer to all of downstream classification and segmentation tasks, which maintain the signature ability of the pre-training. 

\vspace{-0.7em}
\section*{Acknowledgments}
\vspace{-0.5em}
We are grateful fpr the MIT-IBM Watson AI Lab, in particular John Cohn for generously providing the computing resources necessary to conduct this research. Wang's work is in part supported by the NSF Energy, Power, Control, and Networks (EPCN) program (Award number: 1934755), and by an IBM faculty research award.

\clearpage

{\small
\bibliographystyle{ieee_fullname}
\bibliography{CV_tickets}
}

\clearpage

\appendix

\section{More Technical Details}

\subsection{Pruning Algorithm}
Following the routines in previous LTH \cite{frankle2018the,chen2020lottery} works, the algorithm~\ref{alg:imp} outlines the full iterative magnitude pruning (IMP) procedure.

{
\begin{algorithm}[H]
    \small
    \caption{Iterative Magnitude Pruning (IMP)}
    \begin{algorithmic}[1]
    \State Set the initial mask to $m = 1^{d_1}$, with the pre-training $\theta_p$.
    \Repeat
    \State Train $f(x; m \odot \theta_p, \gamma_p)$ for $t$ epochs with algorithm $\mathcal{A}^\mathcal{T}$, i.e., $\mathcal{A}^\mathcal{T}_{t}(f(x; m \odot \theta_p, \gamma_p))$
    \State Prune $20\%$ of remaining weights in $\mathcal{A}^\mathcal{T}_t(f(x; m \odot \theta_p, \gamma_p))$ and update $m$ accordingly
    \Until{the sparsity of $m$ reaches the desired sparsity level $s$}
    \State Return $f(x; m \odot \theta_p)$.
    \end{algorithmic}
    \label{alg:imp}
\end{algorithm}
}

\subsection{Top-1 Retrieval Accuracy} 

Here we presents the detailed calculation of top-1 retrieval accuracy for self-supervised pretraining tasks, including simCLR \cite{chen2020simple} and MoCo \cite{he2020momentum}. Given a batch of data with $n$ samples, $\{z_1,\cdots,z_n\}$ and $\{z'_1,\cdots,z'_n\}$ donates the feature representations from the two branches of simCLR or MoCo models. $z_i$ and $z'_i$ are computed from the same input sample with different data augmentations.

For each $z_i$, we calculate the cosine similarity between $z_i$ and other representations and obtain $\mathcal{D}_i=\{d(z_i,z)|z\in\{z_j,z'_j\}^n_{j=1}/ \{z_i\}\}$, where $d(\cdot,\cdot)$ is the cosine similarity measurement. If $\mathrm{argmax}_{z}\mathcal{D}_i=z'_i$, it suggests the top-1 retrieval is corrected. In the same way, we perform a similar retrieval process for $z'_i$ and $\mathcal{D}'_i$. The concrete calculation formulation of top-1 retrieval accuracy is depicted as follows:
{\begin{align}
    \begin{array}{ll}
     \frac{\sum_{i=1}^n\left[\mathbb{I}(\mathrm{argmax}_{z}\mathcal{D}_i=z'_i)+\mathbb{I}(\mathrm{argmax}_{z}\mathcal{D}'_i=z_i)\right]}{2\times n} \times 100\%,
    \end{array} \label{eq:top1acc}
\end{align}}
where $\mathbb{I}(\cdot)$ is the indicator function.

\section{More Experimental Results}

\subsection{YOLOv4 Detection Results with Other Metrics}

\begin{figure}[t] 
\centering
\includegraphics[width=1\linewidth]{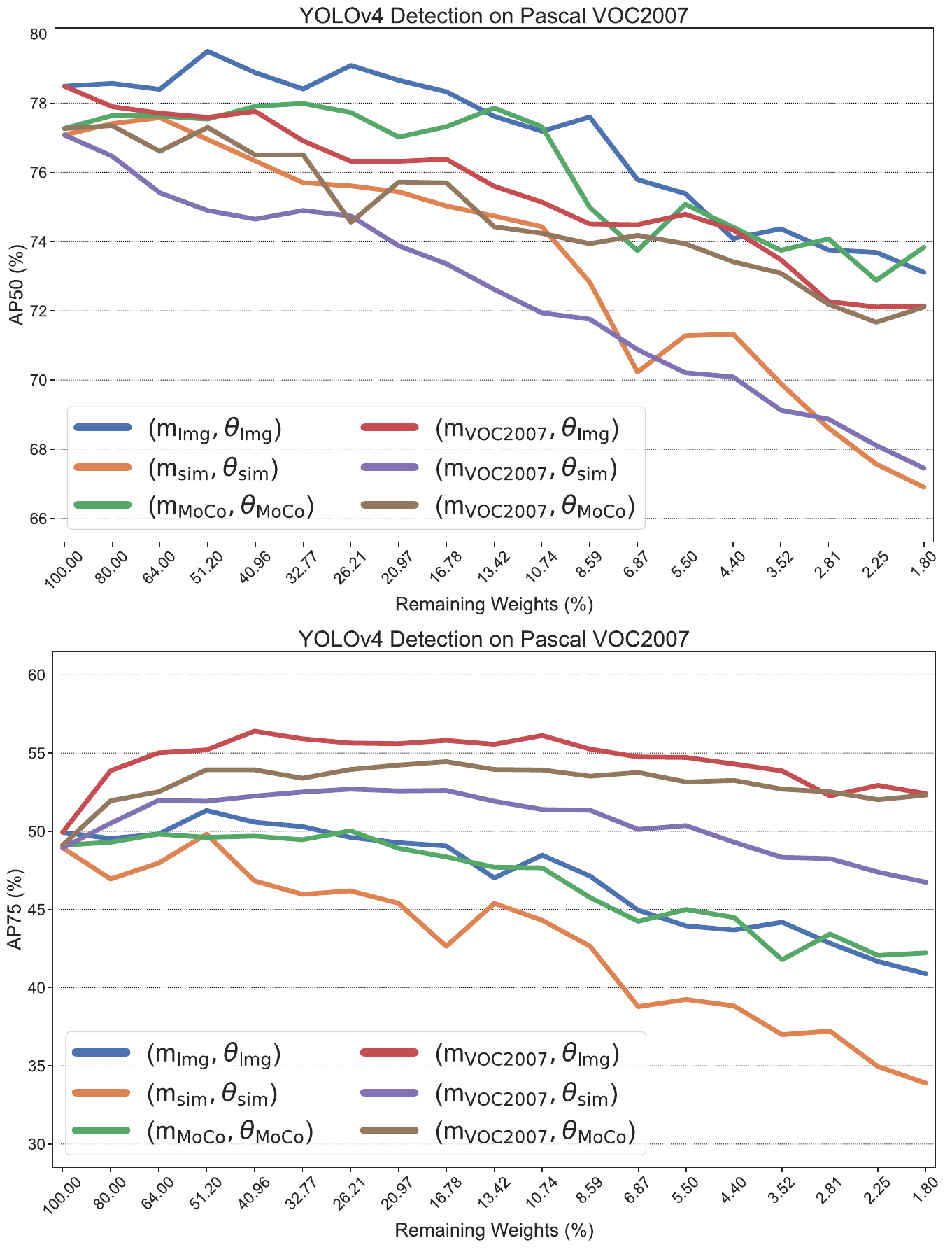}
\caption{Performance (AP$_{50}$ and AP$_{75}$) of IMP subnetworks with a range of sparsity from $0.00\%$ to $98.20\%$ on the downstream tasks. Subnetworks ($m_{\mathrm{VOC2007}},\theta_{p}$), $\theta_p\in\{\theta_{\mathrm{Img}},\theta_{\mathrm{sim}},\theta_{\mathrm{MoCo}}\}$ are identified on the detection task with pre-trained weights $\theta_p$.}
\label{fig:supp_detect}
\end{figure}

In this section, we report the other two evaluation metrics, i.e.,AP$_{50}$ and AP$_{75}$, for YOLOv4 detection experiments. As shown in Figure~\ref{fig:supp_detect}, similar observations can be drawn that there are subnetworks $f(x;m_{\mathcal{P}}\odot\theta_{\mathrm{p}},\cdot)$ capable of transferring to the detection task successfully (i.e., without performance degradation compared with full unpruned models) at the ($83.22\%$,$89.26\%$,$36.00\%$) and ($73.79\%$,$48.80\%$,$79.03\%$) sparsity levels under the AP$_{50}$ and AP$_{75}$ metrics for supervised ImageNet, self-supervised simCLR and MoCo pre-training tasks, respectively.
% the most different observation is that there a subnetwork $f(x;m_{\mathrm{MoCo}}\odot\theta_{\mathrm{MoCo}},\cdot)$ transfer to the detection task successfully (i.e., without performance degradation compared with full unpruned models) at the $59.04\%$ sparsity level under the AP$_{50}$ metric. Other conclusions are consistent with the ones in the main text.

\begin{figure*}[!ht] 
\centering
\includegraphics[width=1\linewidth]{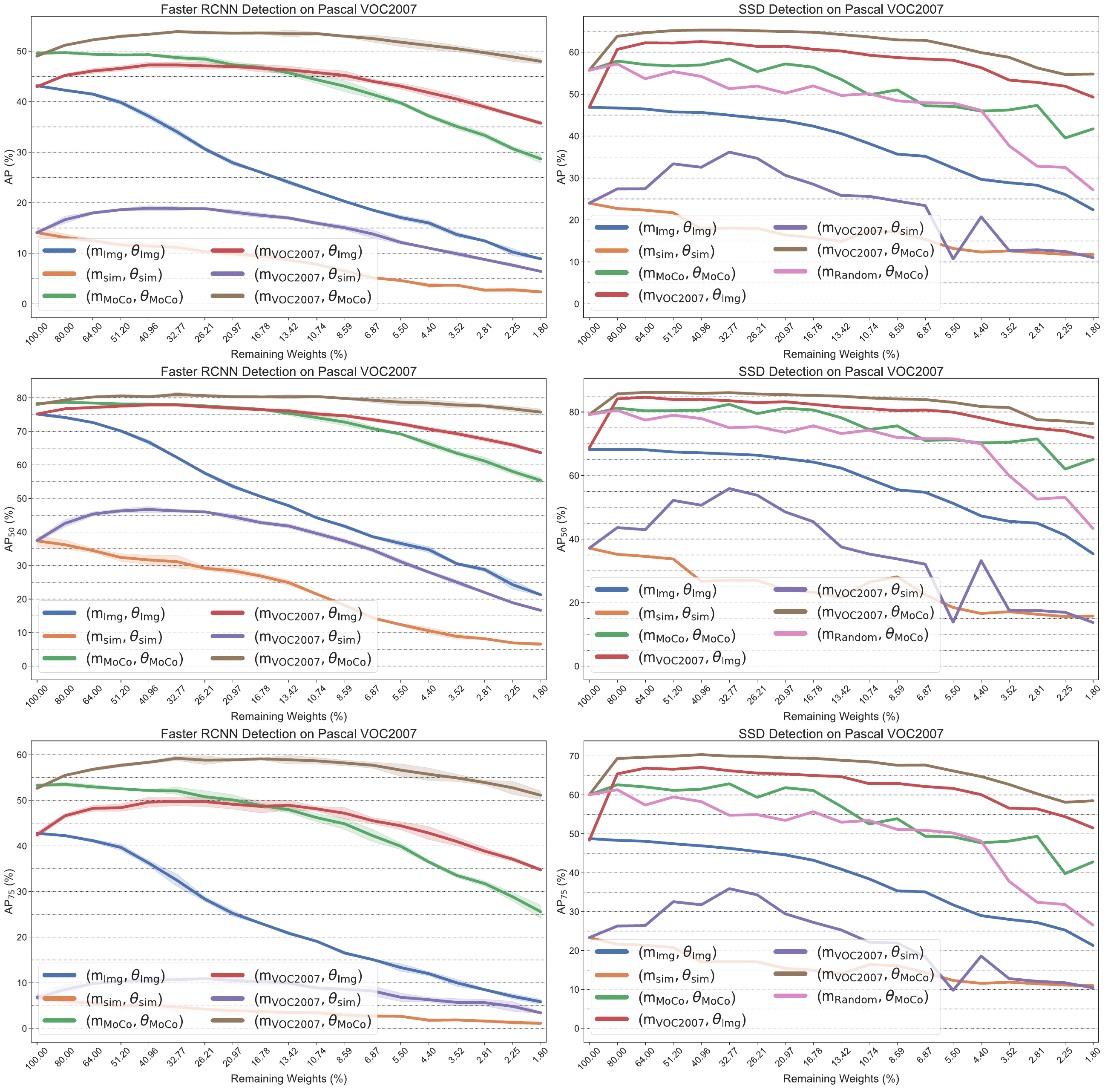}
\caption{Performance (AP, AP$_{50}$ and AP$_{75}$) of IMP subnetworks with a range of sparsity from $0.00\%$ to $98.20\%$ on the downstream tasks. Subnetworks ($m_{\mathrm{VOC2007}},\theta_{p}$), $\theta_p\in\{\theta_{\mathrm{Img}},\theta_{\mathrm{sim}},\theta_{\mathrm{MoCo}}\}$ are identified on the detection task with pre-trained weights $\theta_p$. Results of Faster RCNN with three independent runs and SSD with one run are presented here.}
\label{fig:other_detect}
\end{figure*}

\subsection{Faster RCNN and SSD Detection Results} \label{sec:more_detectors}
In this section, we conduct extra experiments with Faster RCNN \cite{ren2015faster} and SSD \cite{liu2016ssd} on Pascal VOC datasets. Specifically, we train detection models for 24K/120K iterations with a batch size 4/32, a polynomial learning rate (LR) decay (with power 0.9 and initial LR 0.005) / a multi-step learning rate decay (with initial LR 0.001 amd $\times 0.1$ at the 80K, 110K iterations), SGD optimizer with 0.9 momentum, and 0.0001/0.0005 weight decay for Faster RCNN/SSD detectors, respectively. As shown in~\ref{fig:other_detect}, the most different observation is the winning tickets $f(x;m_{\mathcal{P}}\odot\theta_p,\cdot)$ found on the pre-training tasks are almost no longer matching subnetworks on the detection task with both Faster RCNN and SSD, which incurs performance degradation compared to unpruned dense models $f(x;\theta_p,\cdot)$. There is an exception that the subnetworks $f(x;m_{\mathrm{MoCo}}\odot\theta_{\mathrm{MoCo}},\cdot)$ successfully transfer to the detection task with Faster RCNN at the $59.04\%$ sparsity level only under the AP$_{50}$ metric, and with SSD at the $83.22\%$, $86.58\%$, $83.22\%$ sparsity level under AP, AP$_{50}$, AP$_{75}$ metrics, respectively. Other conclusions are consistent with the ones of YOLOv4 detector in the main text.

We notice that detection results of Faster RCNN and SSD with the simCLR pre-training show inferior and unsatisfactory performances, compared with the reported number in BYOL \cite{grill2020bootstrap}. Although BYOL is implemented with Tensorflow (we use Pytorch) and also has an extra residual block for backbone network, the performance gap is not neglectable. To address this, multiple authors have worked to carefully tune all hyperparameters (learning rate, batch size, training iterations), and thoroughly compared implementation details side-to-side (batch norm, input resolution, etc.). However, we still cannot close the gap. Hence while our results on MoCo and ImageNet are very consistent, we cannot exclude the marginal possibility that simCLR implementation is specifically sensitive to Pytorch versus Tensorflow frameworks (unfortunately, not uncommon) for some reason. Therefore, we put Faster RCNN and SSD detection results with the simCLR pre-training in the appendix as failure cases, and note that it hardly affects any of our main observations/conclusions.

\subsection{Ablation about Larger Pre-training Models}

\begin{figure}[!ht] 
\centering
\includegraphics[width=1\linewidth]{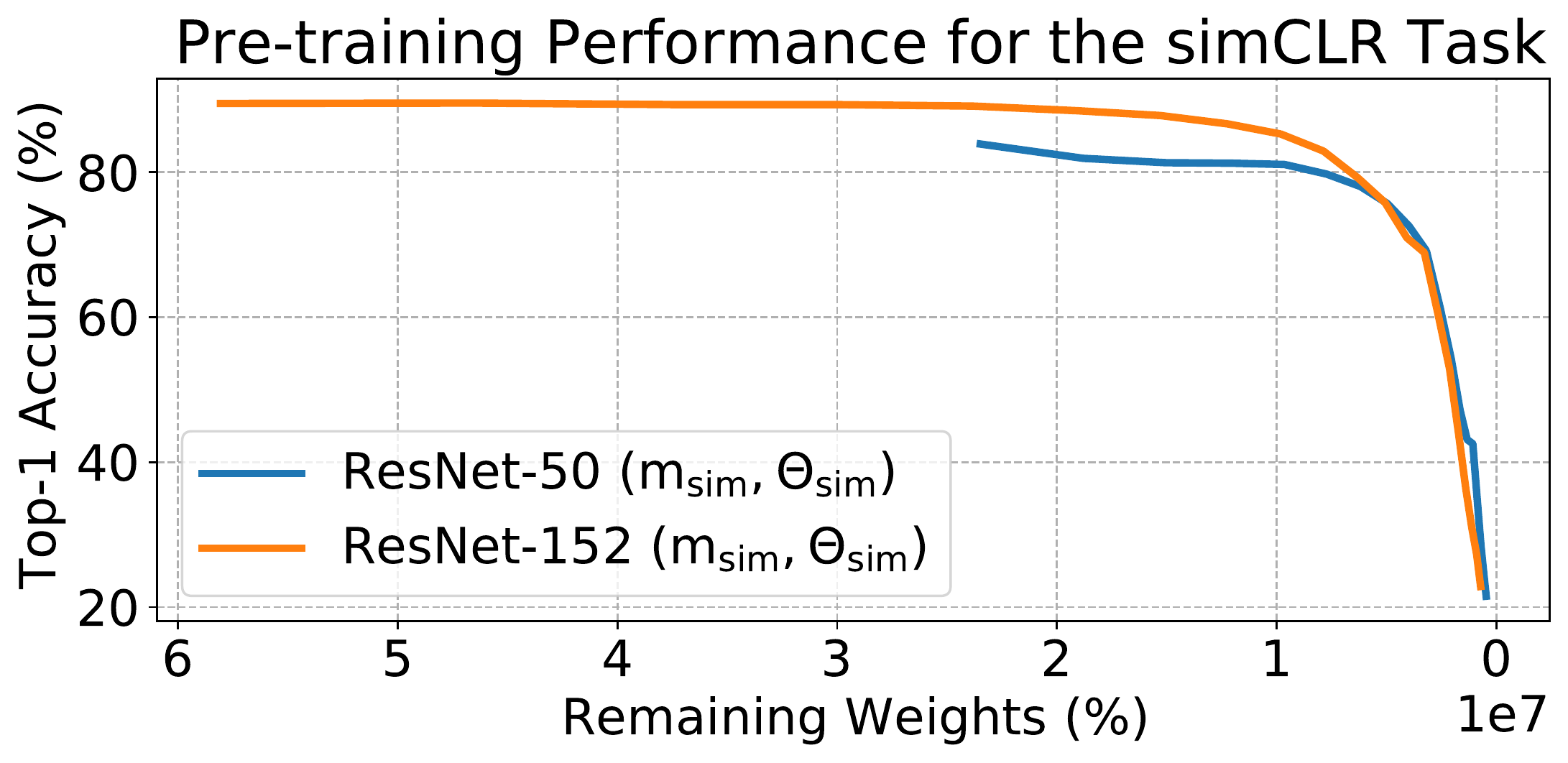}
\caption{Pre-training performance (top-1 retrieval accuracy as defined in Equation~\ref{eq:top1acc}) over the number of remaining weights. Subnetworks are found on the simCLR pre-training task with pre-trained ResNet-50 and ResNet-152 weights.}
\label{fig:preacc}
\end{figure}

Figure~\ref{fig:preacc} collects the pre-training task performance of subnetworks generated from small- and large-scale pre-trained simCLR models. We observe that heavily compressed, large simCLR models (e.g., ResNet-50) obtain superior performance to lightly compressed, small simCLR models (e.g., ResNet-152), which is consistent with \cite{li2020train}. However, subnetworks found on the small-scale pre-trained simCLR model show a slightly better top-1 retrieval accuracy after the sparsity approaches an extreme level.   

\end{document}